\documentclass{article}
\usepackage{spconf,amsmath,graphicx}

\title{Localization of Critical Findings in Chest X-Ray without Local Annotations Using Multi-Instance Learning}
\name{Evan Schwab$^{\dagger}$ \qquad  Andr{\'e} Goo{\ss}en$^{\star}$ \qquad Hrishikesh Deshpande$^{\star}$  \qquad Axel Saalbach$^{\star}$}

\address{$^{\dagger}$Clinical Informatics, Solutions \& Services, Philips Research North America, Cambridge, MA, USA \\
   $^{\star}$Digital Imaging, Philips Research, Hamburg, Germany}
\begin{document}
%
\maketitle
\begin{abstract}
The automatic detection of critical findings in chest X-rays (CXR), such as pneumothorax, is important for assisting radiologists in their clinical workflow like triaging time-sensitive cases and screening for incidental findings. While deep learning (DL) models has become a promising predictive technology with near-human accuracy, they commonly suffer from a lack of explainability, which is an important aspect for clinical deployment of DL models in the highly regulated healthcare industry. For example, localizing critical findings in an image is useful for explaining the predictions of DL classification algorithms.  While there have been a host of joint classification and localization methods for computer vision, the state-of-the-art DL models require locally annotated training data in the form of pixel level labels or bounding box coordinates. In the medical domain, this requires an expensive amount of manual annotation by medical experts for each critical finding. This requirement becomes a major barrier for training models that can rapidly scale to various findings. In this work, we address these shortcomings with an interpretable DL algorithm based on multi-instance learning that jointly classifies and localizes critical findings in CXR without the need for local annotations. We show competitive classification results on three different critical findings (pneumothorax, pneumonia, and pulmonary edema) from three different CXR datasets.
\end{abstract}

\begin{figure}[t]
  \centering
    \includegraphics[width=.9\linewidth, trim={0 0 0 0}, clip]{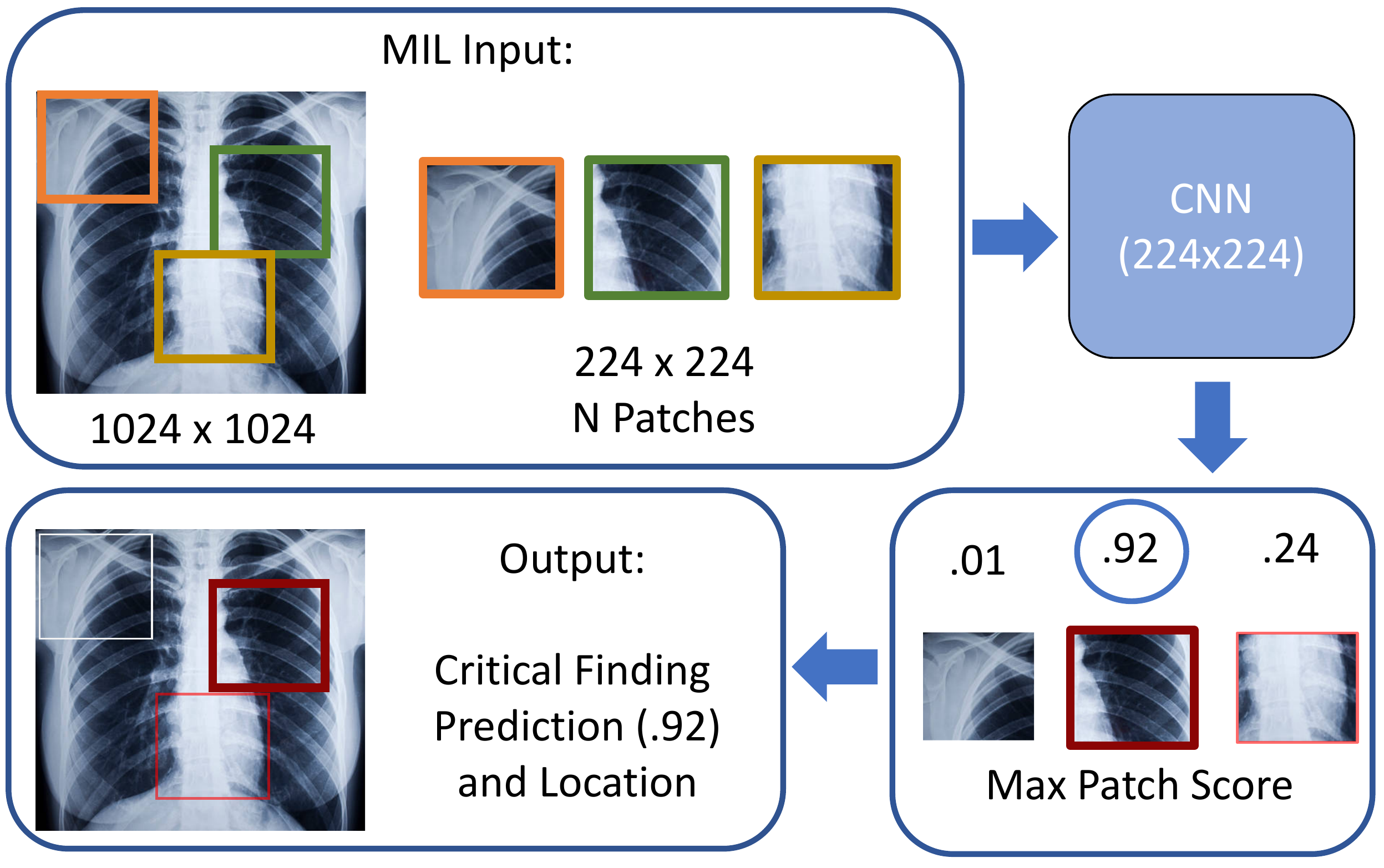}
   \caption{Proposed multi-instance learning network for joint classification and localization of critical findings in CXR.}
   \label{fig:MIL}
   \vspace{-10pt}
\end{figure}

\begin{figure*}[h]
  \centering
    \includegraphics[width=0.18\textwidth, trim={0 0 0 0}, clip]{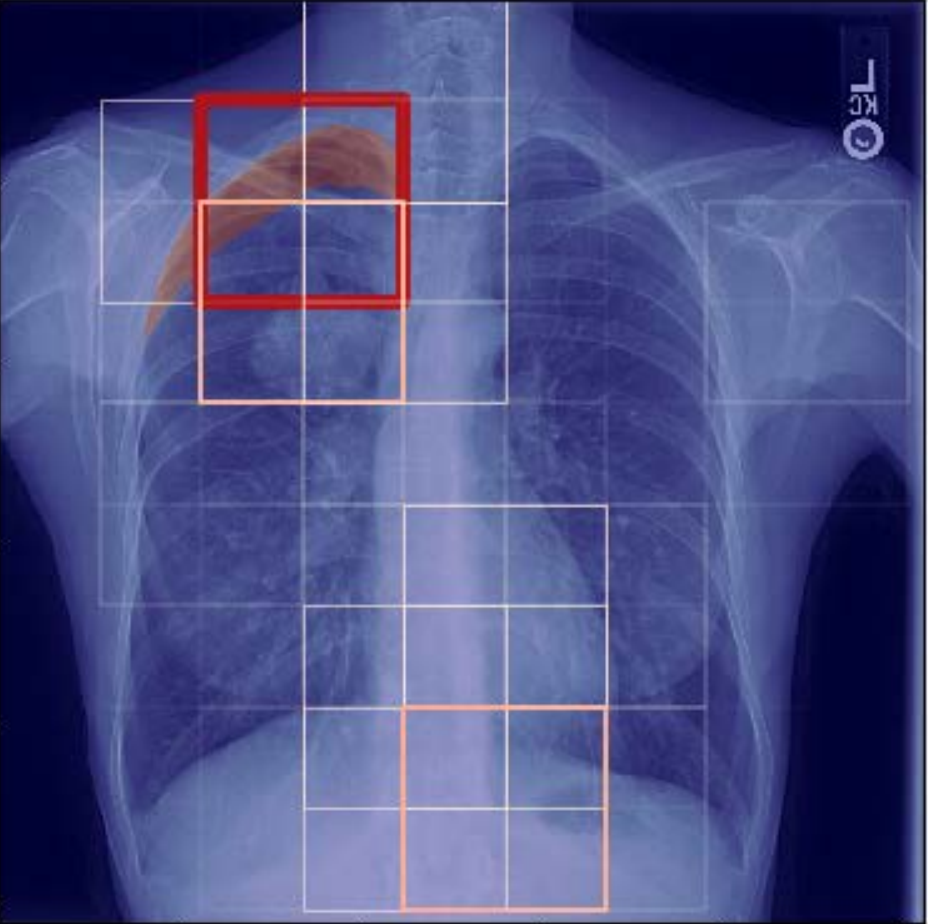}
    \includegraphics[width=0.18\textwidth, trim={0 0 0 0}, clip]{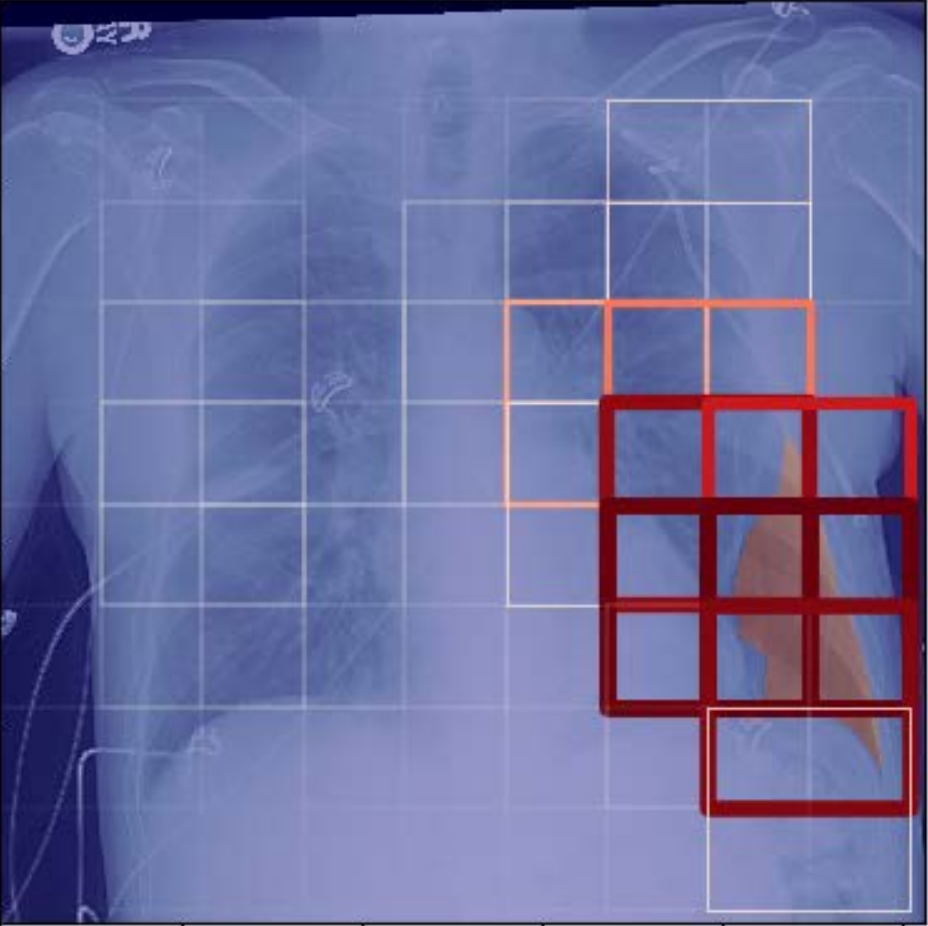}
     \includegraphics[width=0.18\textwidth, trim={0 0 0 0}, clip]{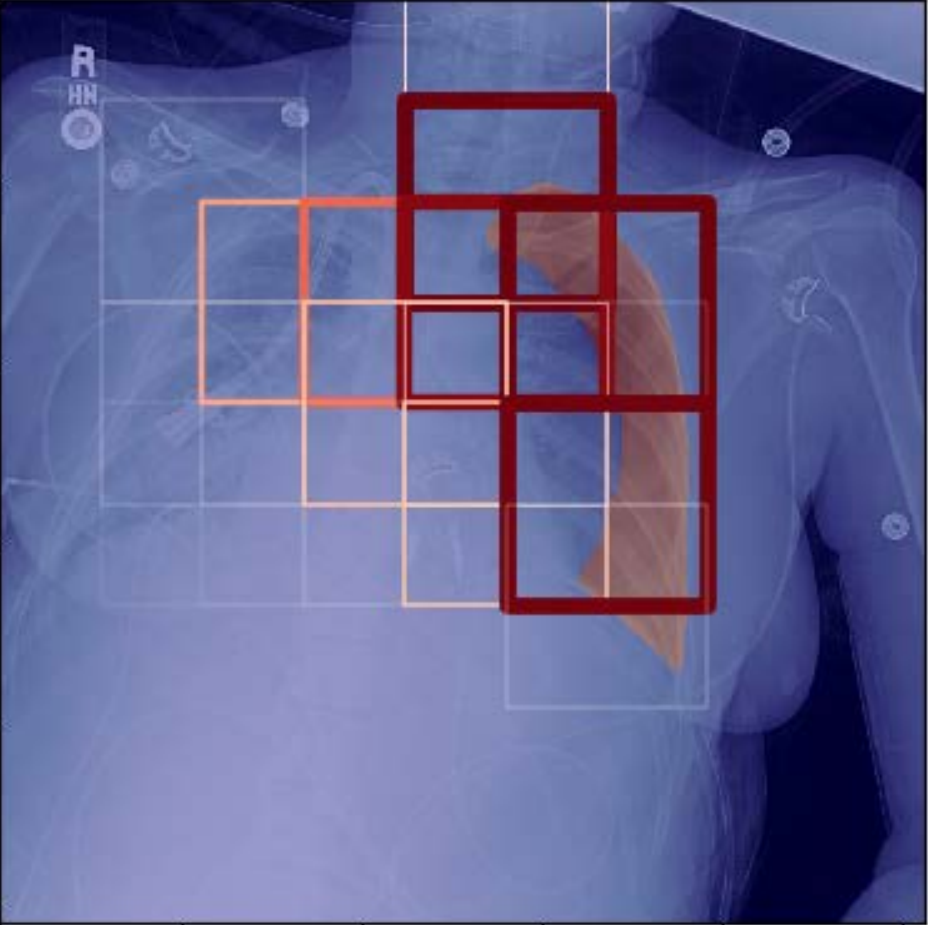}
       \includegraphics[width=0.18\textwidth, trim={0 0 0 0}, clip]{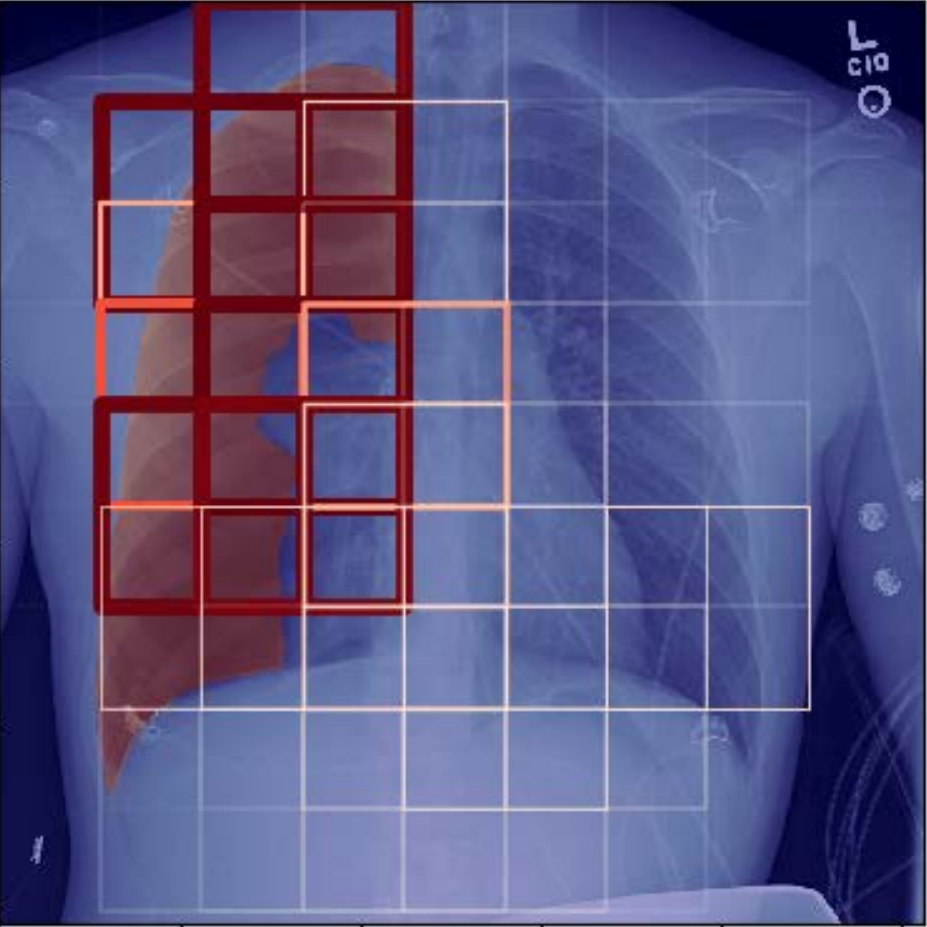}
     \includegraphics[width=0.18\textwidth, trim={0 0 0 0}, clip]{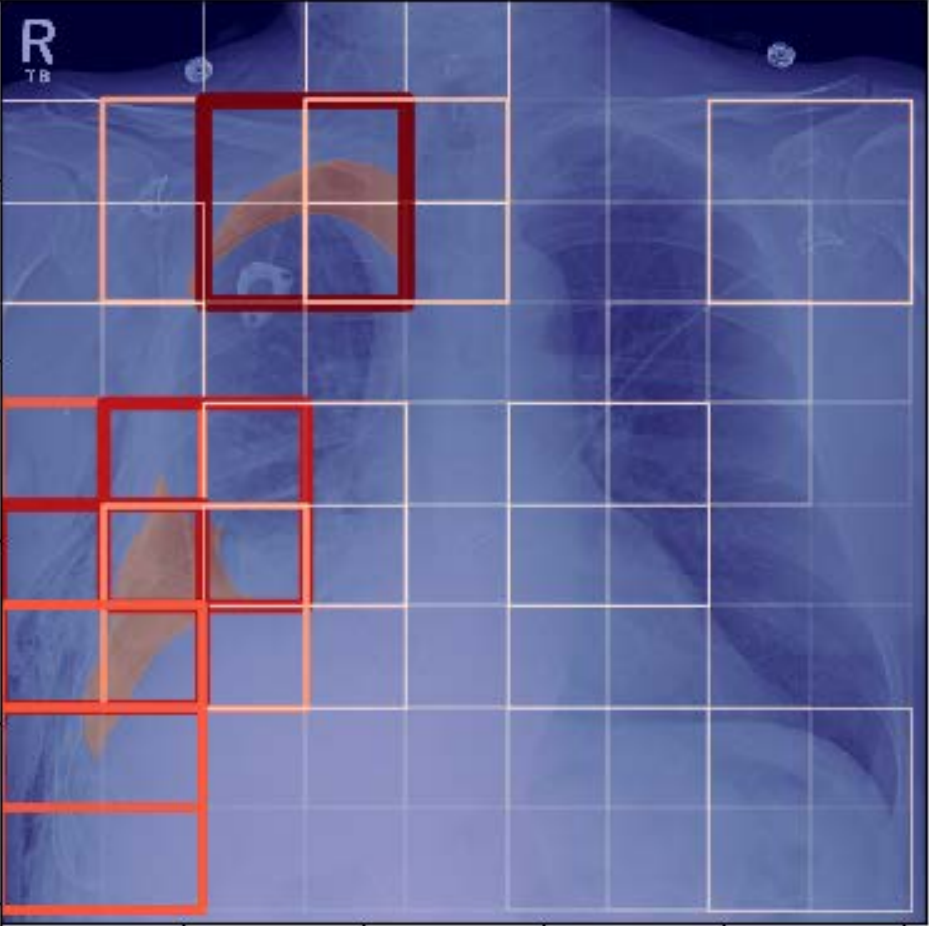}\\
     \includegraphics[width=0.18\textwidth, trim={0 0 0 0}, clip]{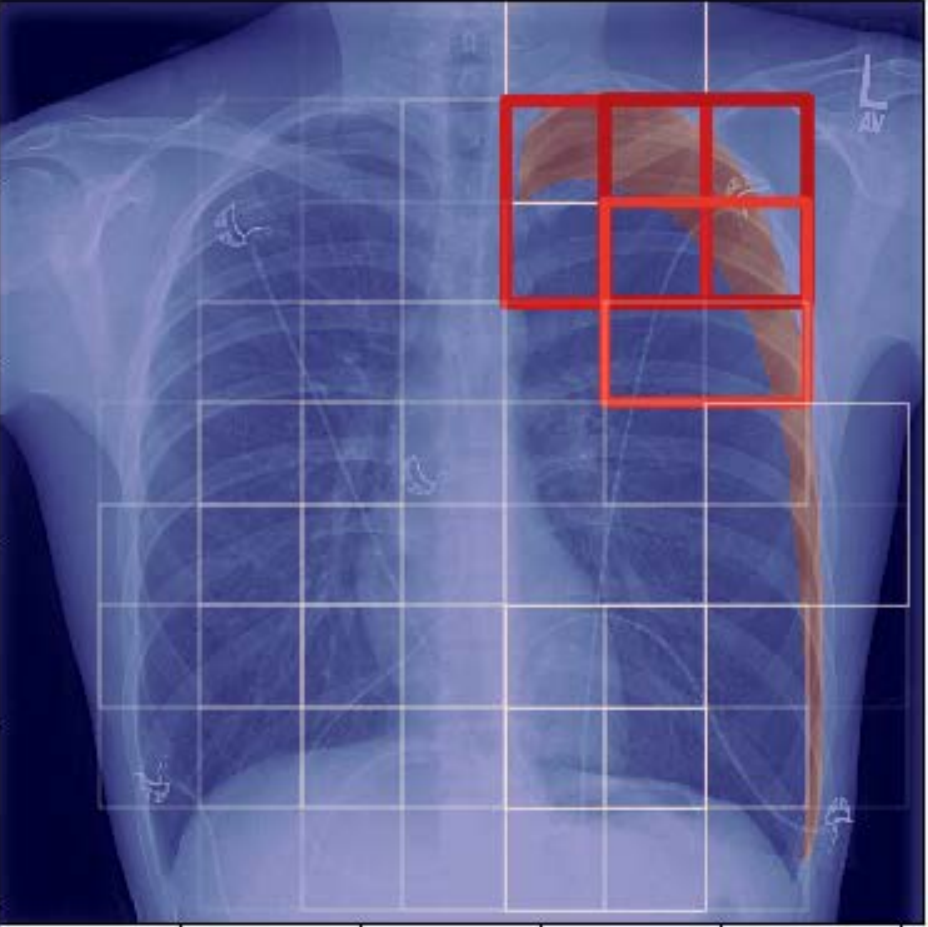}
        \includegraphics[width=0.18\textwidth, trim={0 0 0 0}, clip]{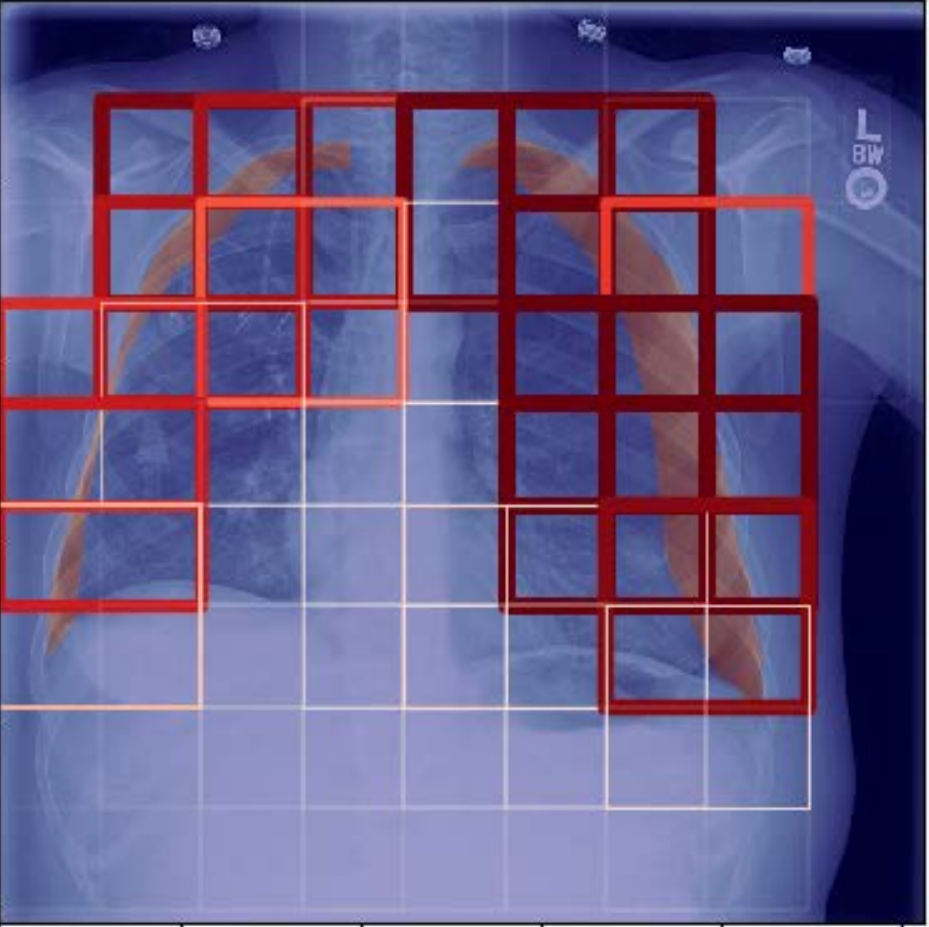}
     \includegraphics[width=0.18\textwidth, trim={0 0 0 0}, clip]{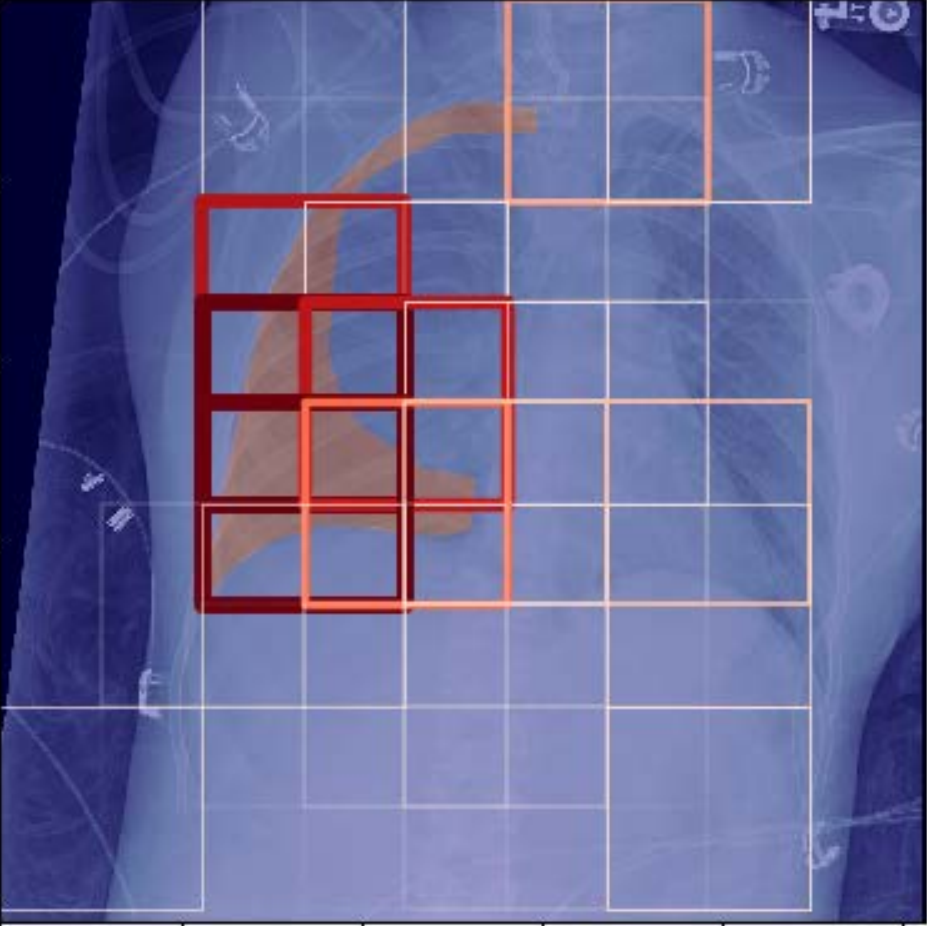}
     \includegraphics[width=0.18\textwidth, trim={0 0 0 0}, clip]{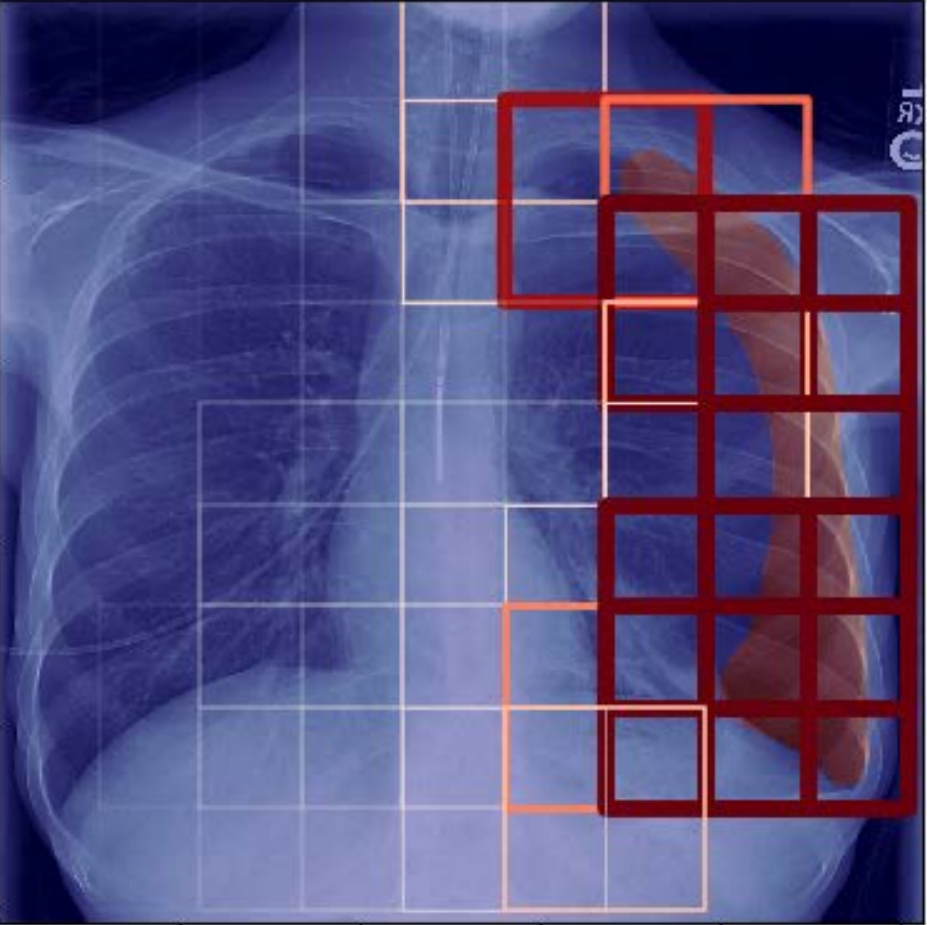}
  \includegraphics[width=0.18\textwidth, trim={0 0 0 0}, clip]{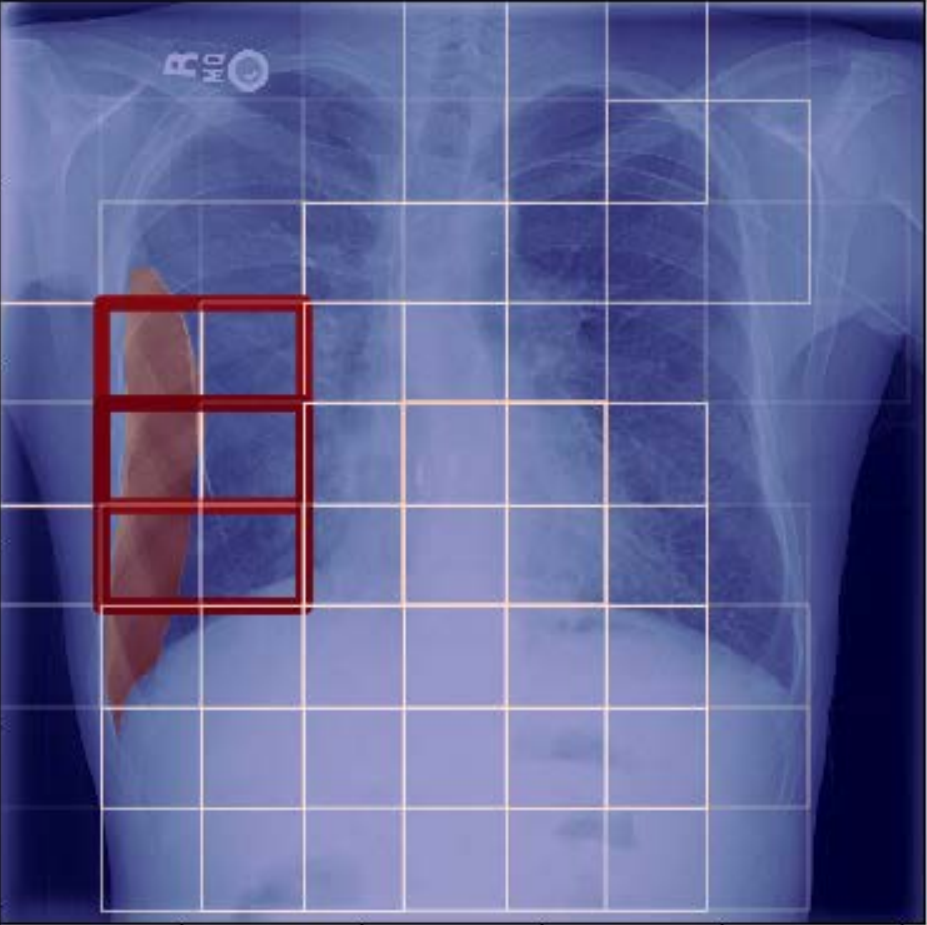}\\
     \includegraphics[width=0.18\textwidth, trim={0 0 0 0}, clip]{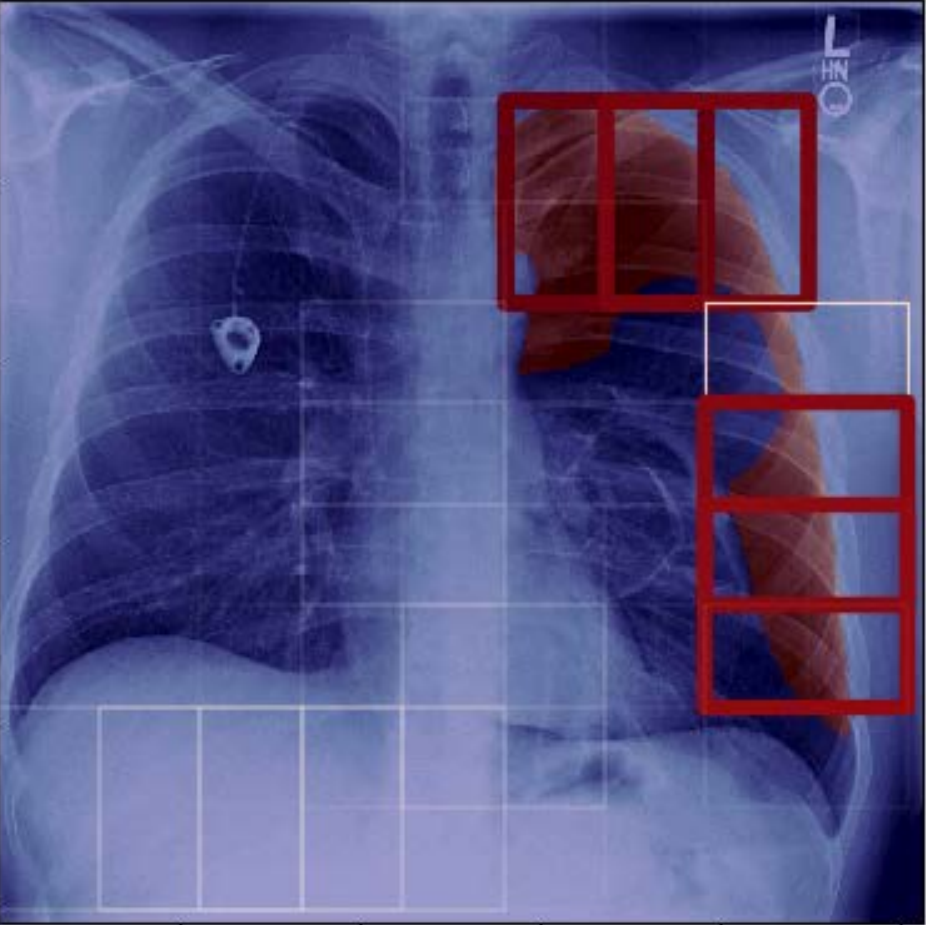}
     \includegraphics[width=0.18\textwidth, trim={0 0 0 0}, clip]{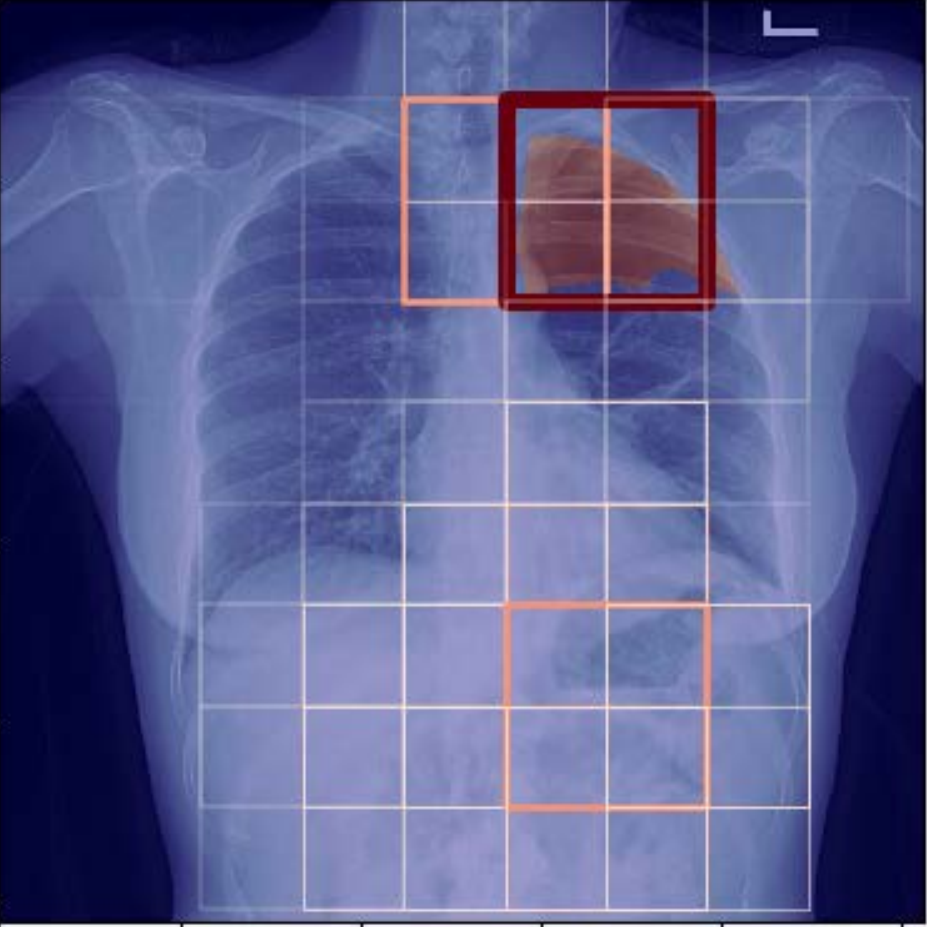}
       \includegraphics[width=0.18\textwidth, trim={0 0 0 0}, clip]{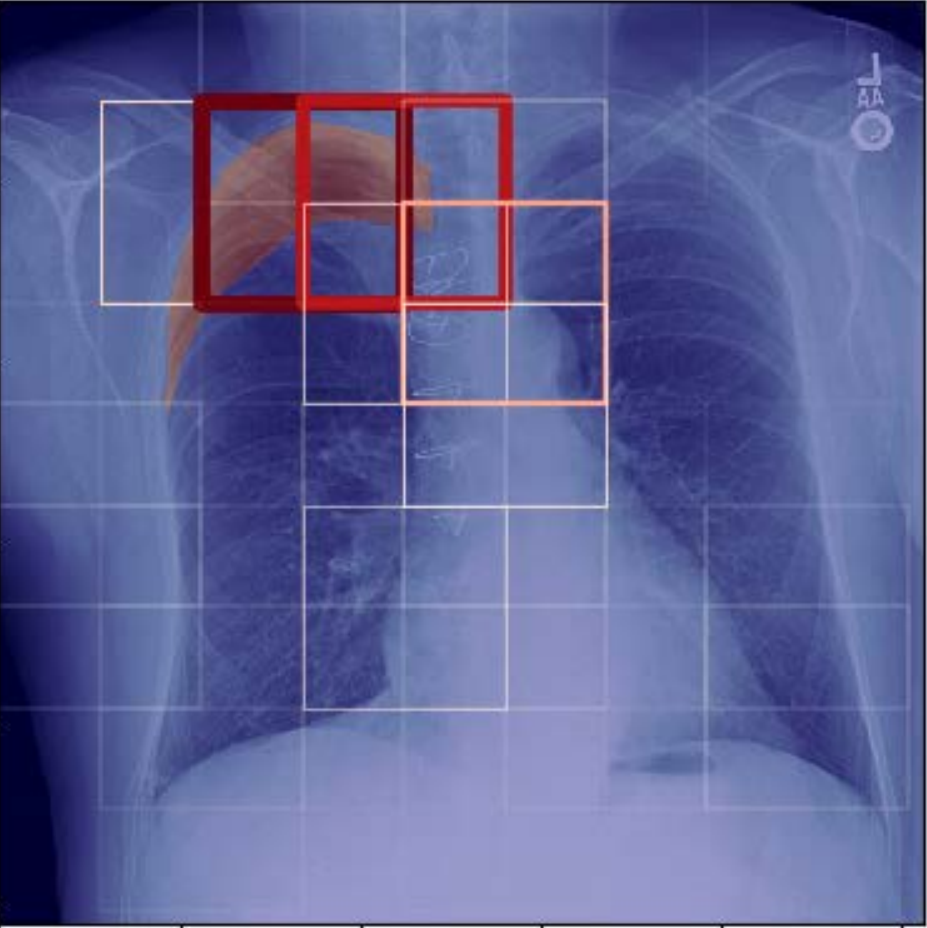}
     \includegraphics[width=0.18\textwidth, trim={0 0 0 0}, clip]{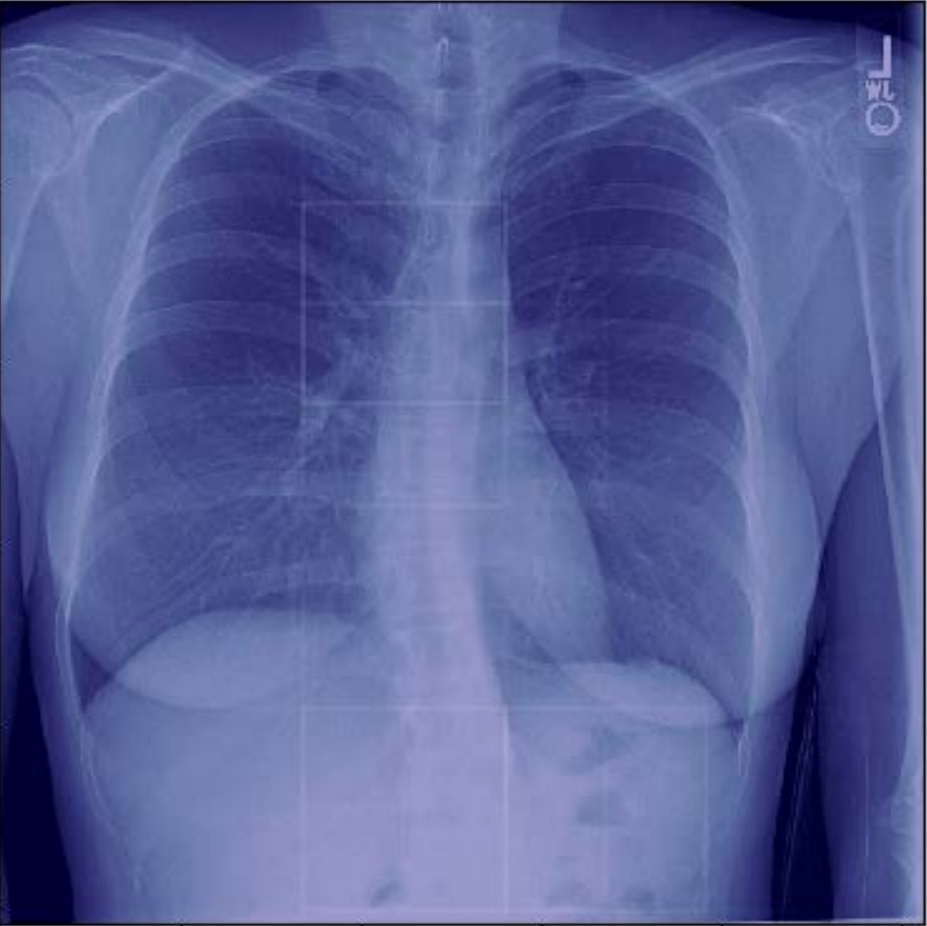}
      \includegraphics[width=0.18\textwidth, trim={0 0 0 0}, clip]{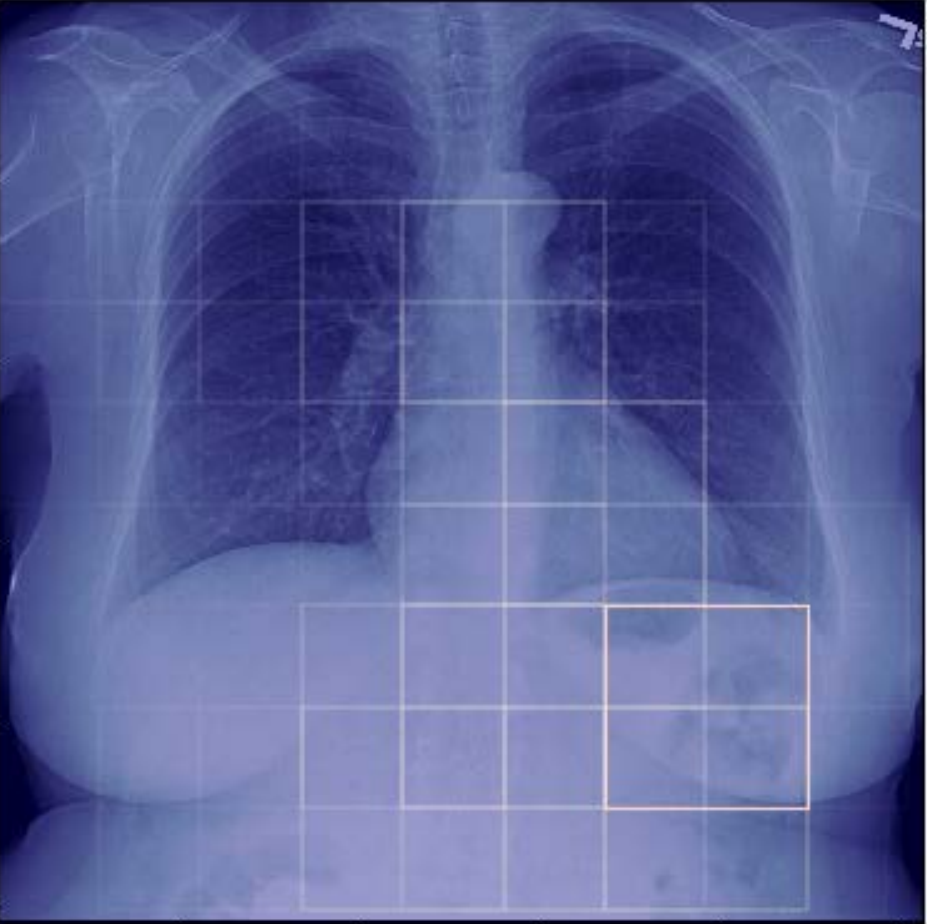}
\caption{MIL localization results for PTX with ground truth segmentations overlayed in red. The patch scores range from 0 (white) to 1 (dark red), with border thickness equal to patch score, i.e. patches close to 0 visually disappear. The score of the image having PTX is equal to max patch score. The last two images are negative for PTX with max patch score close to 0.}
\label{fig:resultsPTX}
\end{figure*}

\section{Introduction}
\label{sec:intro}

The automatic detection of critical findings like pneumothorax (PTX), pneumonia (PNA), or pulmonary edema (PE) in chest X-ray images (CXR) is a highly researched topic with a multitude of clinical use cases \cite{larson2014actionable}. One of the most commonly utilized deep neural networks for image classification is the convolutional neural network (CNN) which takes in an image and outputs an image class prediction. However, while deep neural networks offer near human accuracy for classifying images, they are commonly seen as ``black box" technology whereby the exact reasons for the classifications are hidden within the complexities of the model. In the case of medical images, providing an explanation for a classification prediction of a critical finding is important for clinicians to trust the output of algorithms especially when critical findings are subtle and difficult to diagnose.


To provide such network explanations, saliency methods such as gradient class activation mapping (Grad-CAM) \cite{selvaraju2017grad,wang2017chestx,guan2018diagnose} provide pixel-wise heatmaps indicating the locations in the image that contributed to the class prediction.  While these methods provide explanations with respect to the distribution of network weights for each class, they are determined \textit{after} a class prediction has been made and therefore do not inform the classification through the optimization.  Additionally, the heatmaps are generated with respect to low-resolution filters (eg. $7\!\times\!7$) and projected back to the size of the input image, resulting in sometimes coarse localizations. This is especially worrisome for medical images like CXR which are often acquired in very high resolution (eg. $3000\!\times\!3000$). Commonly used CNN methods will first down-sample CXR to meet the sizes of pre-trained networks from ImageNet (eg. $224\!\times\!224$) which may degrade details for accurate localization.



To provide localization \textit{during} the optimization, object detection \cite{redmon2016you,li2018thoracic} and segmentation \cite{ronneberger2015u} algorithms are widely used to localize objects in an image by predicting regional bounding boxes or pixel-wise classes. To do so, these algorithms require local annotations of bounding box coordinates or pixel-level labels to be predicted in combination with the image class.  For natural images local labels can be feasibly crowd-sourced, but in the medical domain, annotating critical findings in medical images requires expertly trained radiologists and the subtlety and variability of each finding make it a highly time-consuming task.

In this work, we address both of these current shortcomings, lack of interpretability and a need for expensive local annotations, with a joint classification and localization algorithm whereby the image-level classification is guided explicitly by the localization and without using local annotations. We introduce the core methodology in Sec.~\ref{sec:methods}, experimental results in Sec.~\ref{sec:experiments}, and conclude in Sec.~\ref{sec:conclusion}.


\section{Methods}
\label{sec:methods}
The core methodology in this work is called multi-instance learning (MIL) \cite{yan2016multi,zhu2017deep,xu2014deep,kraus2016classifying} whereby data is broken into a set of parts or \textit{instances} which are collectively analyzed to understand the local aspects that give the data its class label. For our application, we define each instance as an image patch. Then, for CXR (binary) classification, the goal of MIL is to predict the label of each patch as containing a critical finding (positive) or not (negative). This is considered a \textit{weakly} supervised problem because the images have ground truth labels but the individual patches do not. However, we know that a negative CXR will contain only negative patches and a positive CXR will contain at least one positive patch. Therefore, MIL uses this knowledge to learn which patches in a positive image are similar to those in a negative image, leaving the dissimilar patch(es) to be the reason for the positive image label, thereby localizing the critical finding.

To classify patches, each patch is input into a CNN with an output of a patch score between 0 and 1 of the probability of containing a critical finding. Since we do not have patch labels in training, MIL uses a mechanism to relate the patch scores with the image labels. For MIL, the most fundamental function is to take the maximum score over all the patches and set this equal to the image score used in the loss function. Then, the optimization suppresses negative patches towards 0 and maximizes positive patches towards 1, thereby simultaneously classifying positive and negative images and identifying the positive patches responsible for the image classification.

Fig.~\ref{fig:MIL} shows an overview of our MIL algorithm applied to CXR. Our MIL network consists of three main components: 1) division of images into patches, 2) a CNN which produces a probabilistic class score for each patch, and 3) a final max layer taken over all patches in an image which relates the patch scores to the image-level label in a loss function.  

In our implementation, we first standardize our CXR to $1024\!\times\!1024$ and divide each image into set of overlapping patches of size suitable for a CNN model pre-trained on ImageNet. For example, using VGG16 with input size $224\!\times\!224$ we use stride $112\!=\!224/2$, resulting in 64 overlapping patches per image. We use batch size 64 to restrict the max to be taken over all patches in a single image. We use the binary cross-entropy loss function between max patch score (ie. image score) and image labels.The outputs of our algorithm are: 1) an image class prediction equal to the maximum patch score prediction and 2) a set of patch scores providing a patch-level probabilistic localization of the critical finding.

We used SGD with Nesterov acceleration, momentum of 0.9, learning rate of $1e^{-5}$ and decay of $1e^{-6}$ for 250 epochs. We fine-tuned the pre-trained VGG16 after freezing the first 15 layers. 
We utilized augmentation during training, consisting of a random sequence of flipping, scaling, translation and rotation applied to the images before division into patches.

\section{Experiments}
\label{sec:experiments}

We provide binary classification and localization results for three different critical findings from three CXR datasets (See Table~\ref{tab:data}): University of Washington Medical Center (UWMC)\footnote{We acknowledge Drs. Cross (UWMC) and Mabotuwana (Philips) for data acquisition/curation approved by the institutional review board.}, the 2018 RSNA Kaggle Competition\footnote{www.kaggle.com/c/rsna-pneumonia-detection-challenge}, and MIMIC-CXR \cite{johnson2019mimic} from Beth Israel Deaconess Medical Center, in collaboration with MIT. The UWMC and Kaggle have ground truth segmentations and bounding boxes, respectively, which we use only for evaluating correctness visually and not within training. The MIMIC-CXR data has no ground truth annotation. We use $\sim\!1000$ CXR images in each dataset with a 80/20 training/validation split.

\begin{table}[t]
\centering
\begin{tabular}{llll}
\hline
CXR Dataset  & Critical Finding & \#Pos/Neg & Annot. \\
\hline
UWMC     & Pneumothorax  & 437/566 & Seg.   \\
RSNA/Kaggle       & Pneumonia  & 500/500  & B. Box   \\
MIMIC-CXR    & Pulmonary Edema & 500/500 & -    \\
\hline
\end{tabular}
\caption{CXR data with type of local annotations available.}
\vspace{-10pt}
\label{tab:data}
\end{table}


We report the Area Under the Curve (AUC) of the validation receiver operator curve. Using VGG16, we achieved AUCs of 0.89, 0.84, and 0.82 for PTX, PNA, and PE, respectively. As an additional experiment for PTX in UWMC (see \cite{GoossenMIDL2019} for experimental details), we compared our MIL against two additional state-of-the-art classification methods: 1) a modified ResNet-50 CNN with increased field of view ($448\times 448$) pre-trained on NIH ChestX-ray14 dataset \cite{rajpurkar2017chexnet} and 2) a fully convolutional (FCN) segmentation-based method which requires pixel-level labels during training. For fair comparison, the FCN and our MIL methods employed the same pre-trained ResNet-50 backbone architecture. Performing 5-fold cross validation, the average AUCs on the validation set were 0.96 (CNN), 0.92 (FCN), and 0.93 (MIL). The MIL outperforms the FCN and is competitive with the CNN in terms of image classification, while adding localization.

\begin{figure}[t]
  \centering  
     \includegraphics[width=0.15\textwidth, trim={0 0 0 0}, clip]{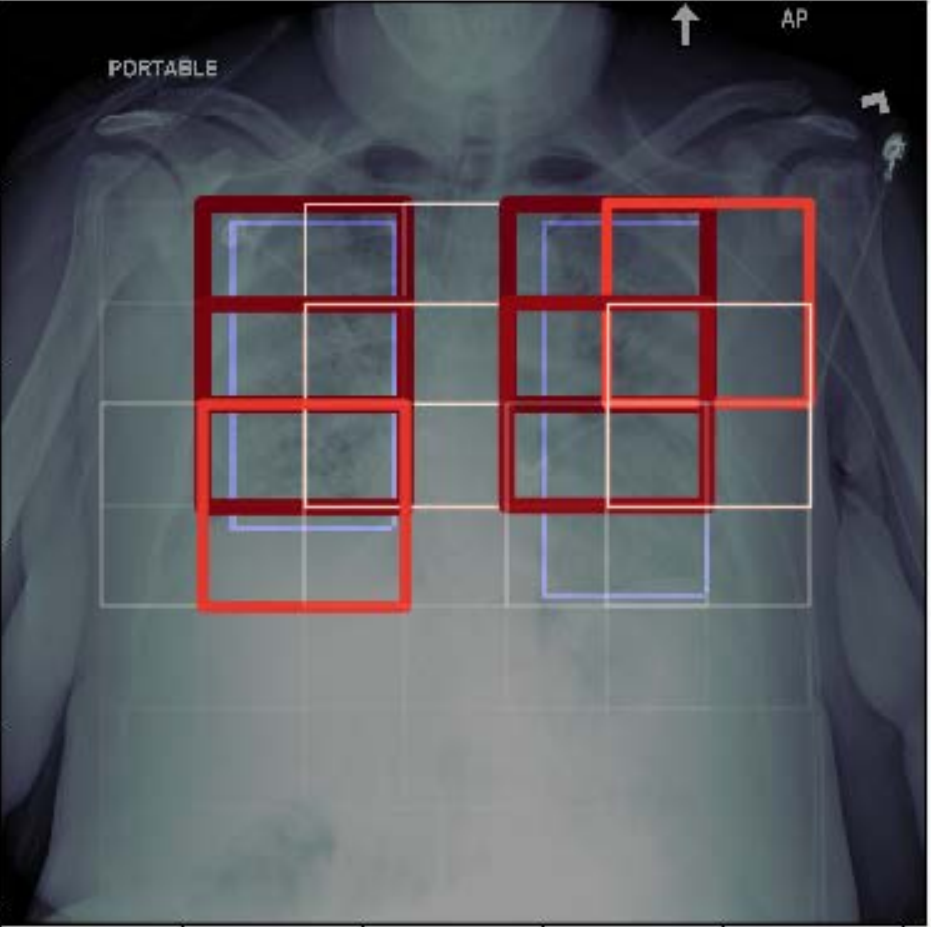}
     \includegraphics[width=0.15\textwidth, trim={0 0 0 0}, clip]{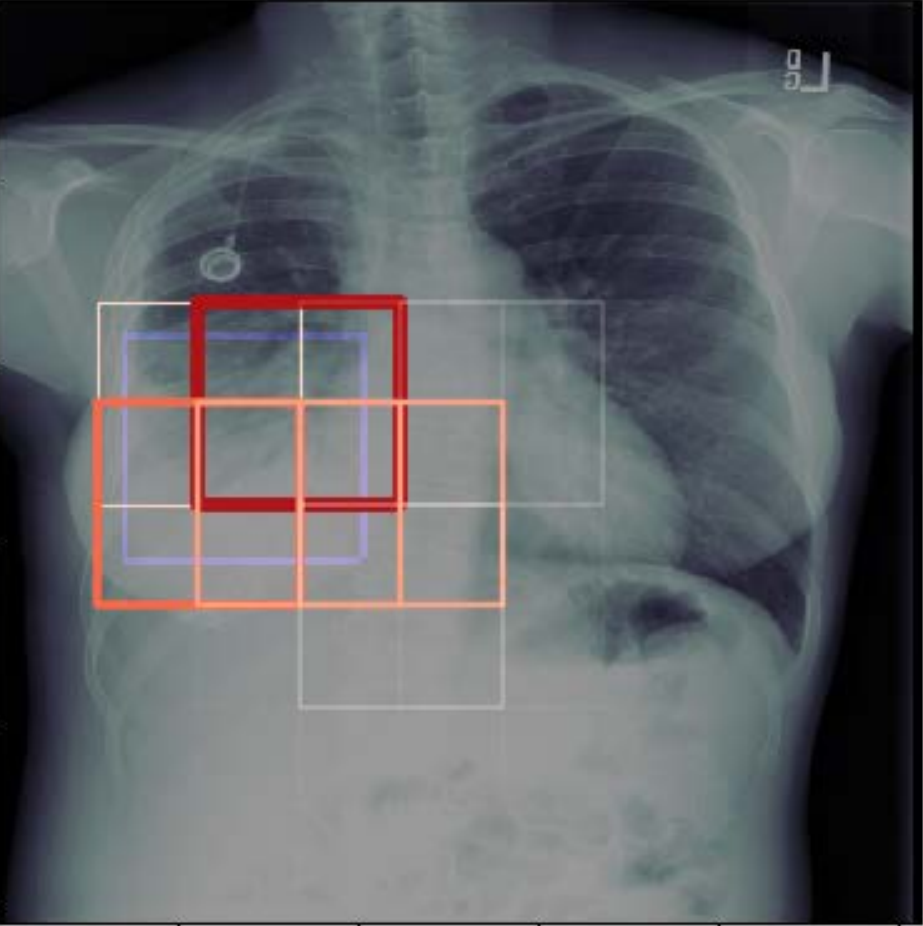}
     \includegraphics[width=0.15\textwidth, trim={0 0 0 0}, clip]{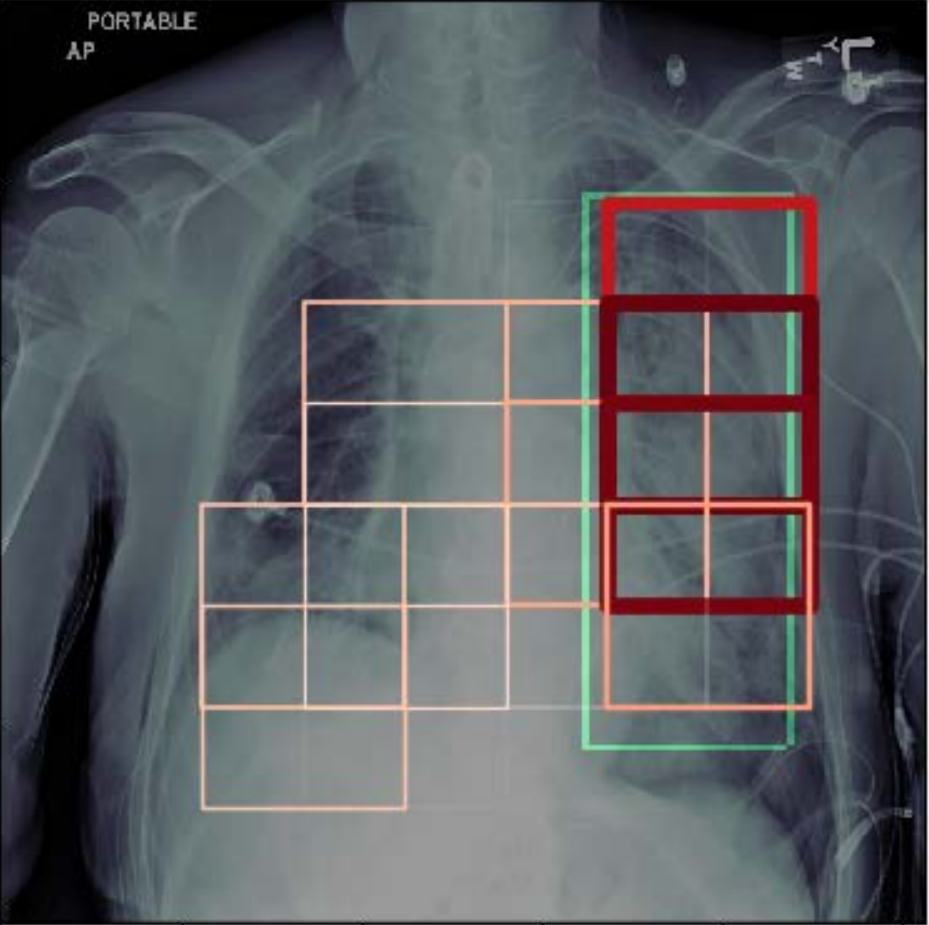}\\
      \includegraphics[width=0.15\textwidth, trim={0 0 0 0}, clip]{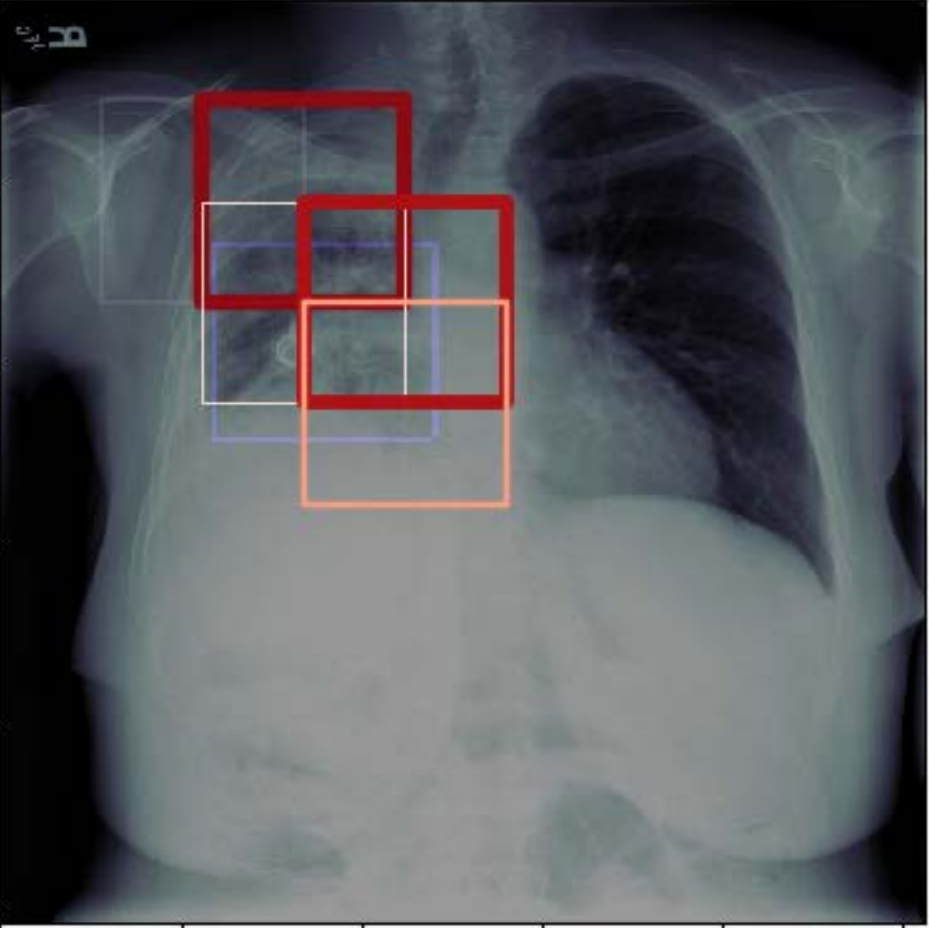}
     \includegraphics[width=0.15\textwidth, trim={0 0 0 0}, clip]{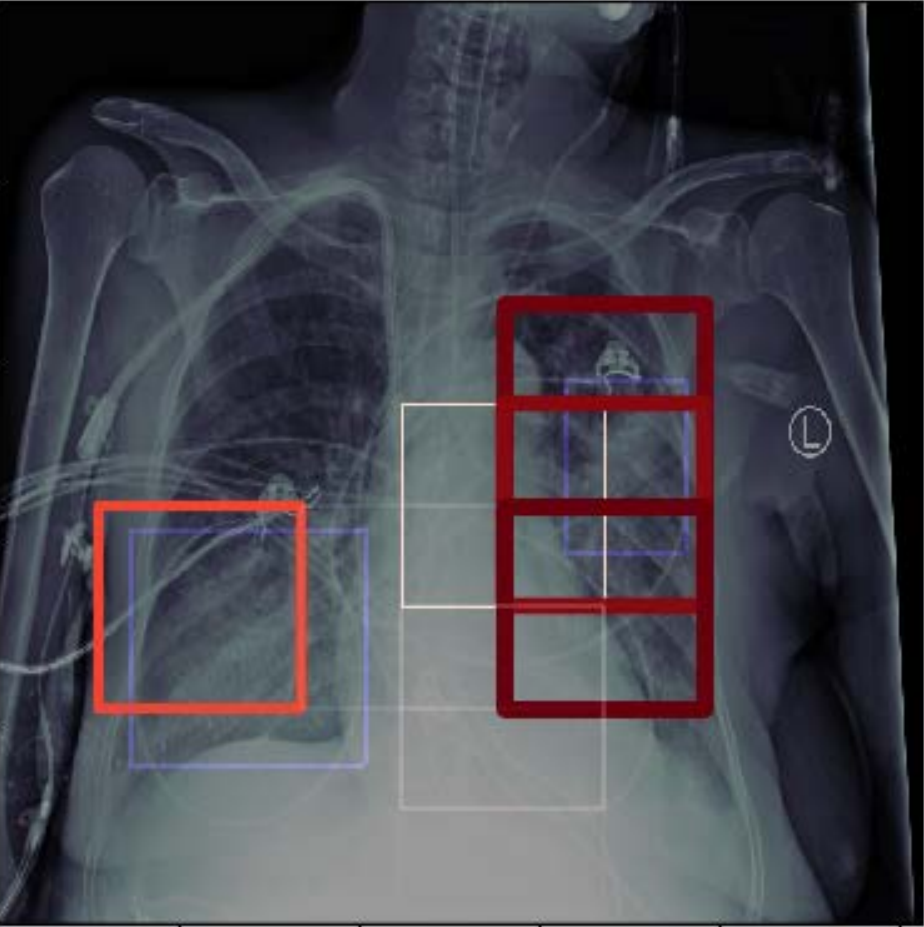}
      \includegraphics[width=0.15\textwidth, trim={0 0 0 0}, clip]{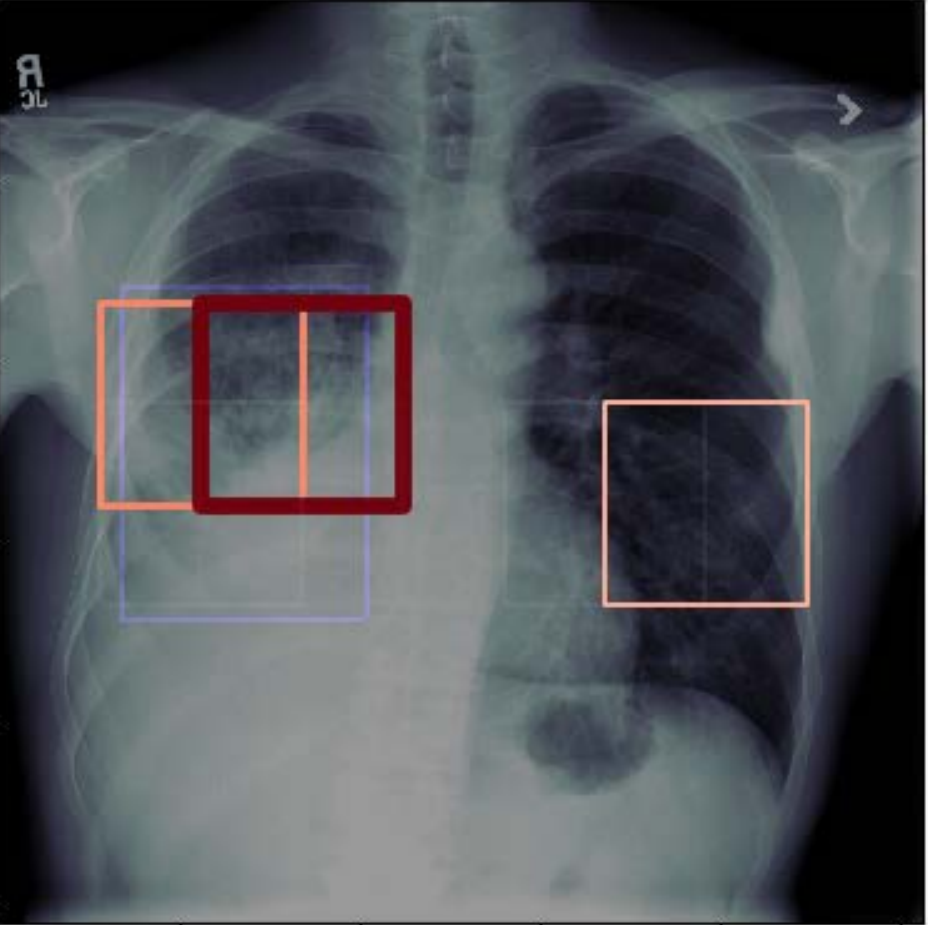}\\
    \includegraphics[width=0.15\textwidth, trim={0 0 0 0}, clip]{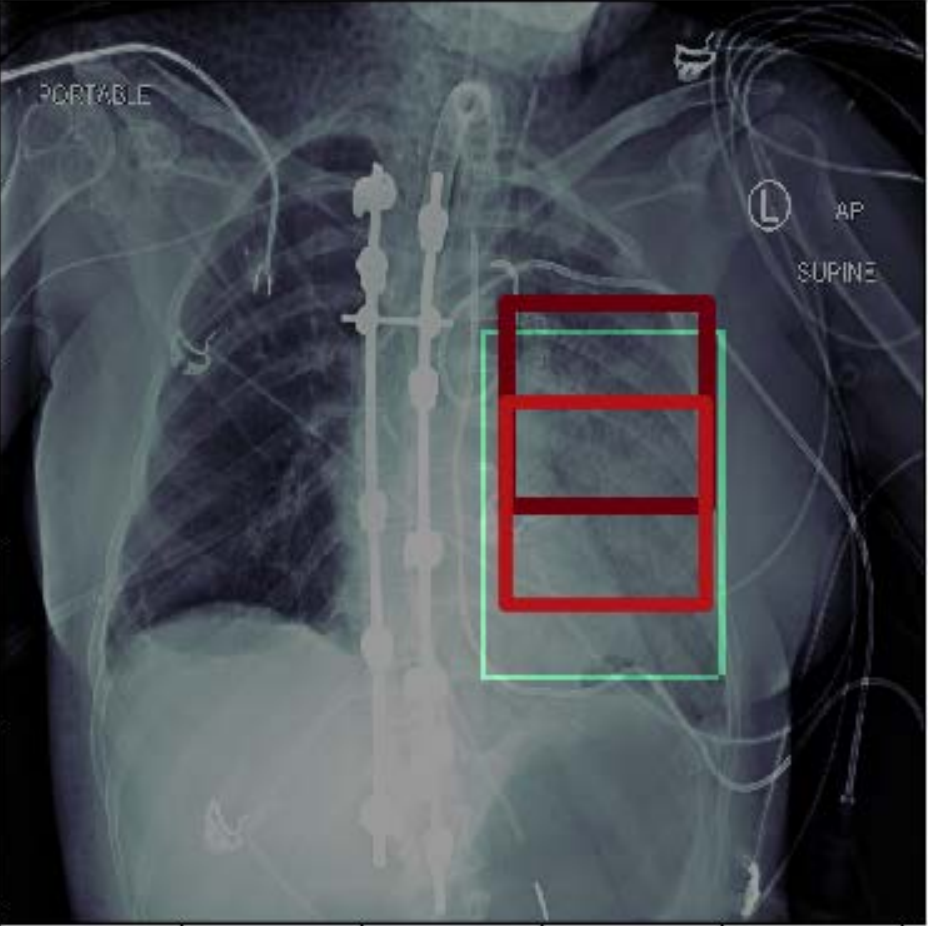}
     \includegraphics[width=0.15\textwidth, trim={0 0 0 0}, clip]{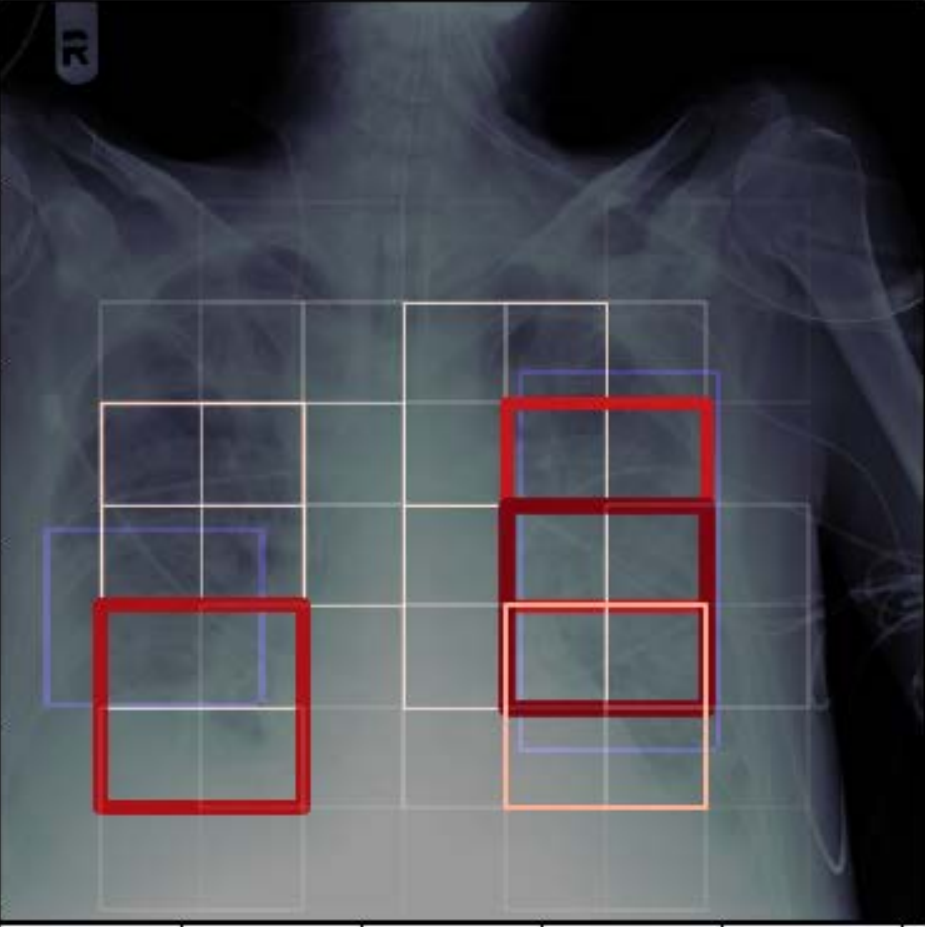}
     \includegraphics[width=0.15\textwidth, trim={0 0 0 0}, clip]{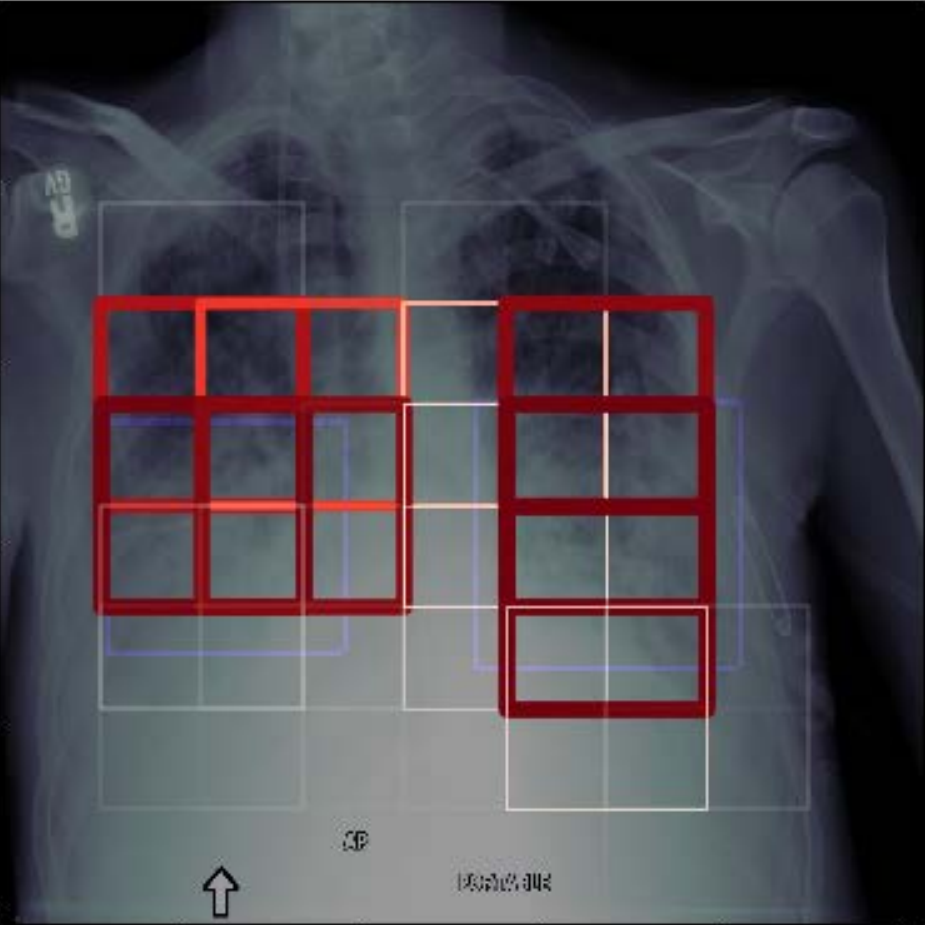}
 \caption{MIL localization results for PNA with ground truth bounding boxes (green or blue).}
      \label{fig:resultsPNA}
\end{figure}

\begin{figure}
          \vspace{-15pt}
  \centering  
     \includegraphics[width=0.15\textwidth, trim={0 0 0 0}, clip]{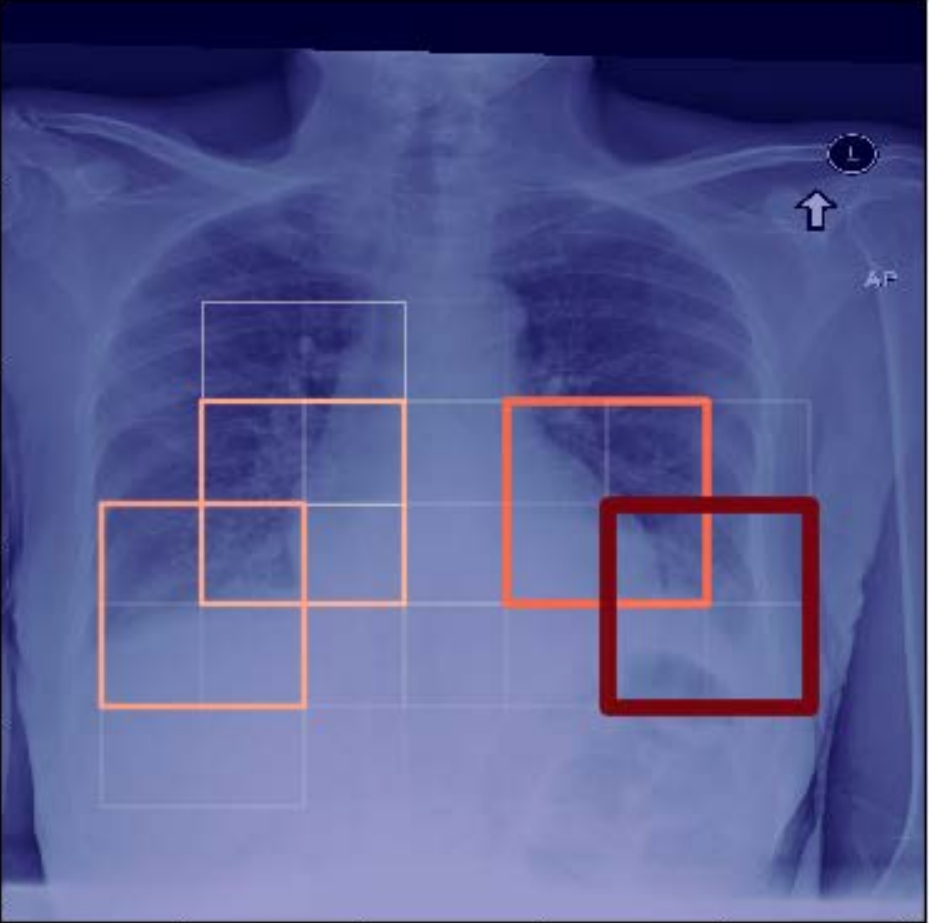}
     \includegraphics[width=0.15\textwidth, trim={0 0 0 0}, clip]{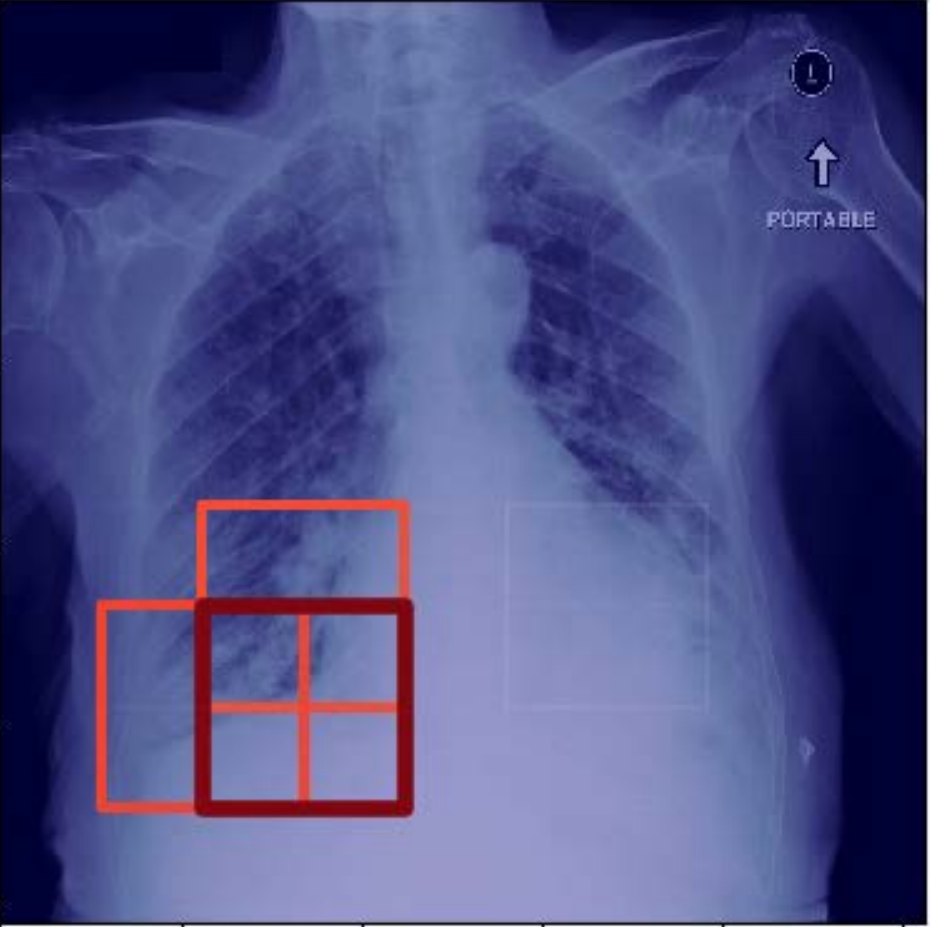}
     \includegraphics[width=0.15\textwidth, trim={0 0 0 0}, clip]{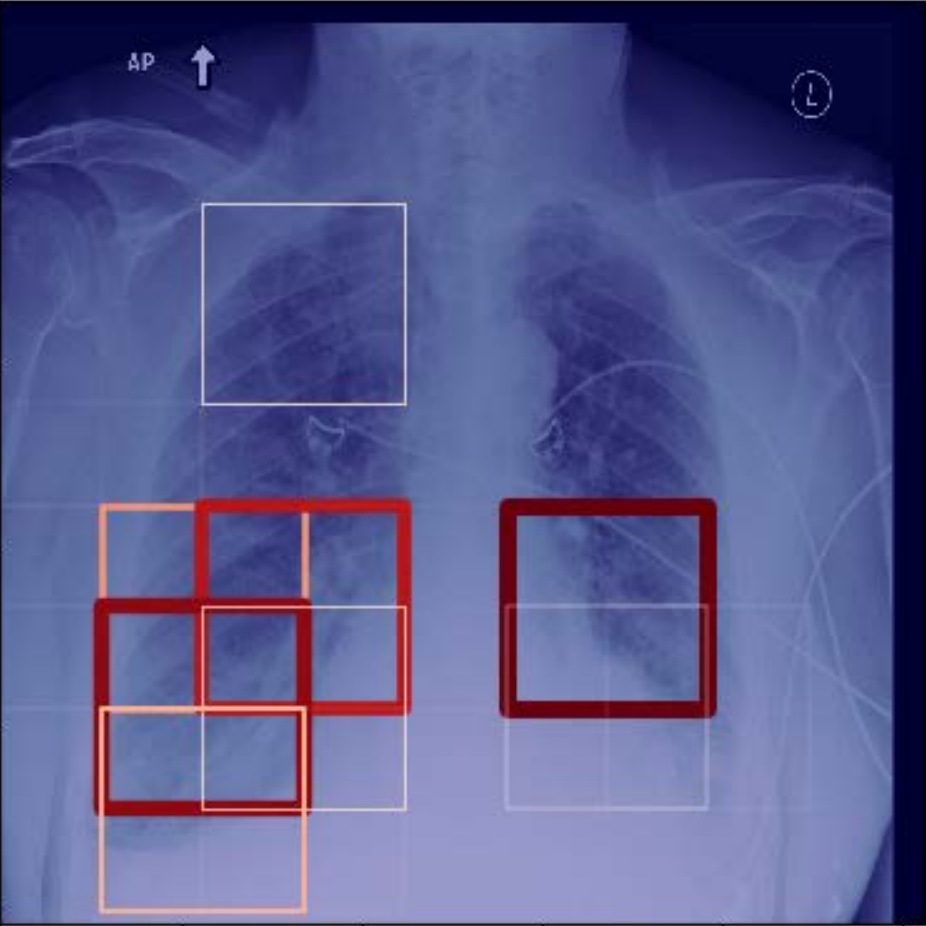}\\
   \includegraphics[width=0.15\textwidth, trim={0 0 0 0}, clip]{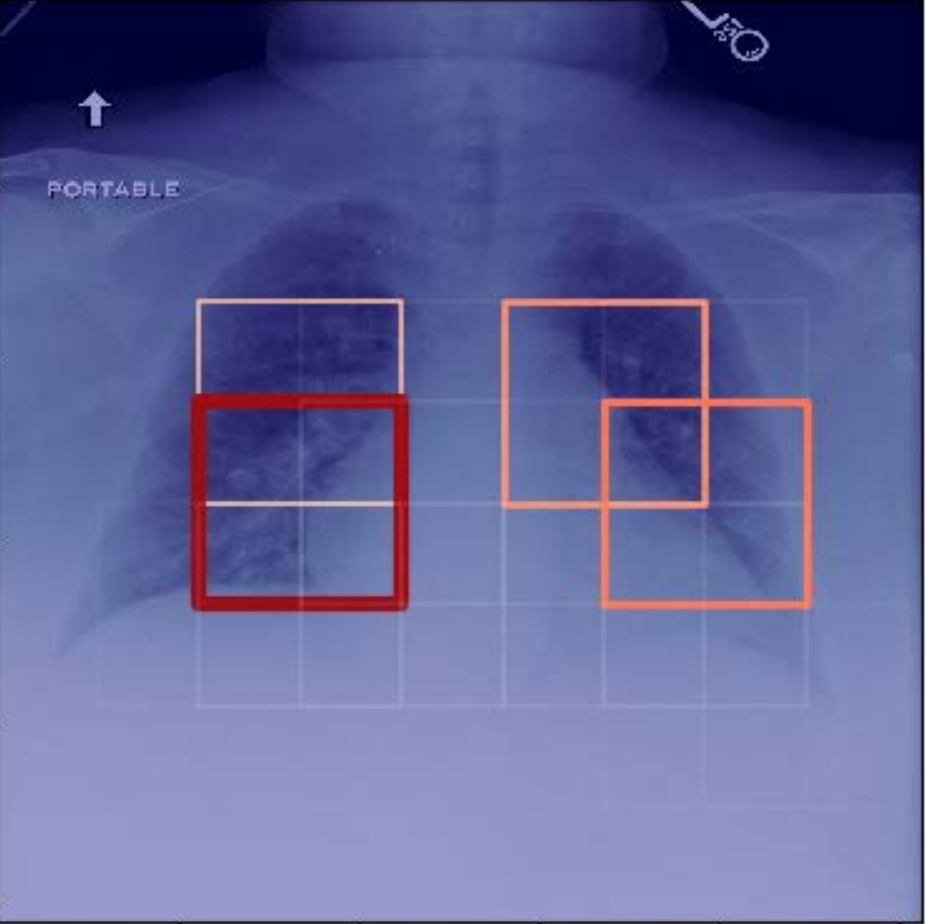}
     \includegraphics[width=0.15\textwidth, trim={0 0 0 0}, clip]{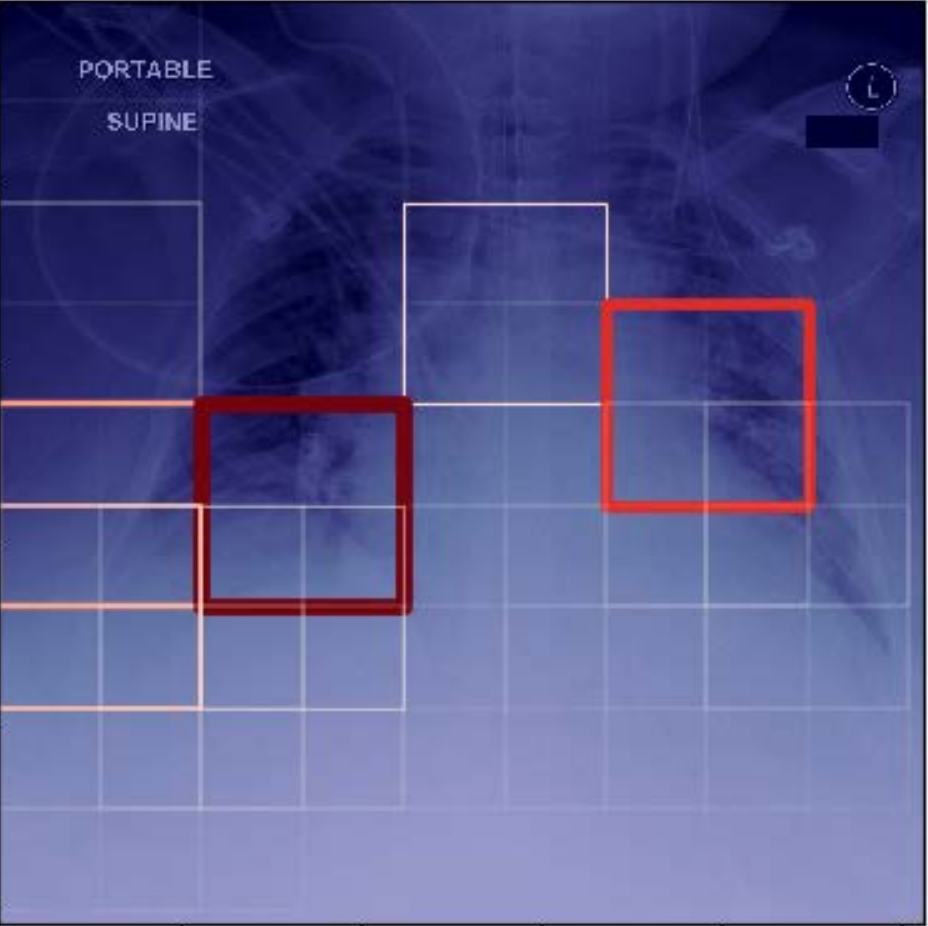}
     \includegraphics[width=0.15\textwidth, trim={0 0 0 0}, clip]{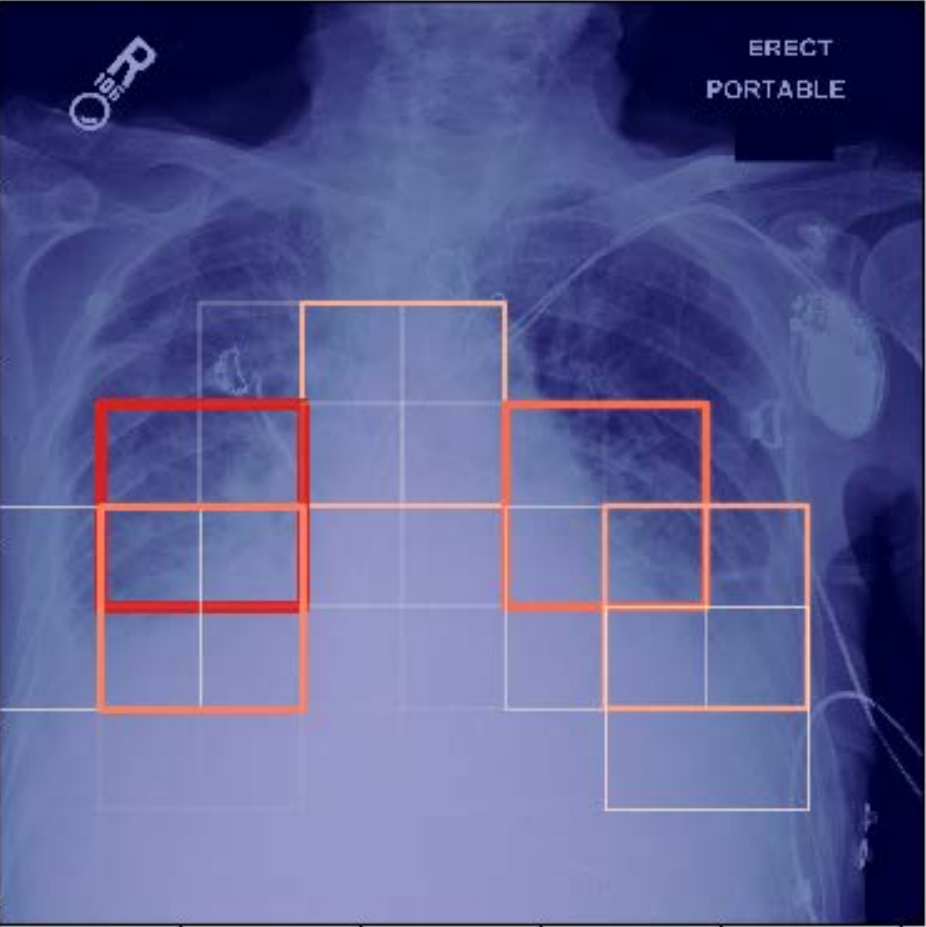}
        \includegraphics[width=0.15\textwidth, trim={0 0 0 0}, clip]{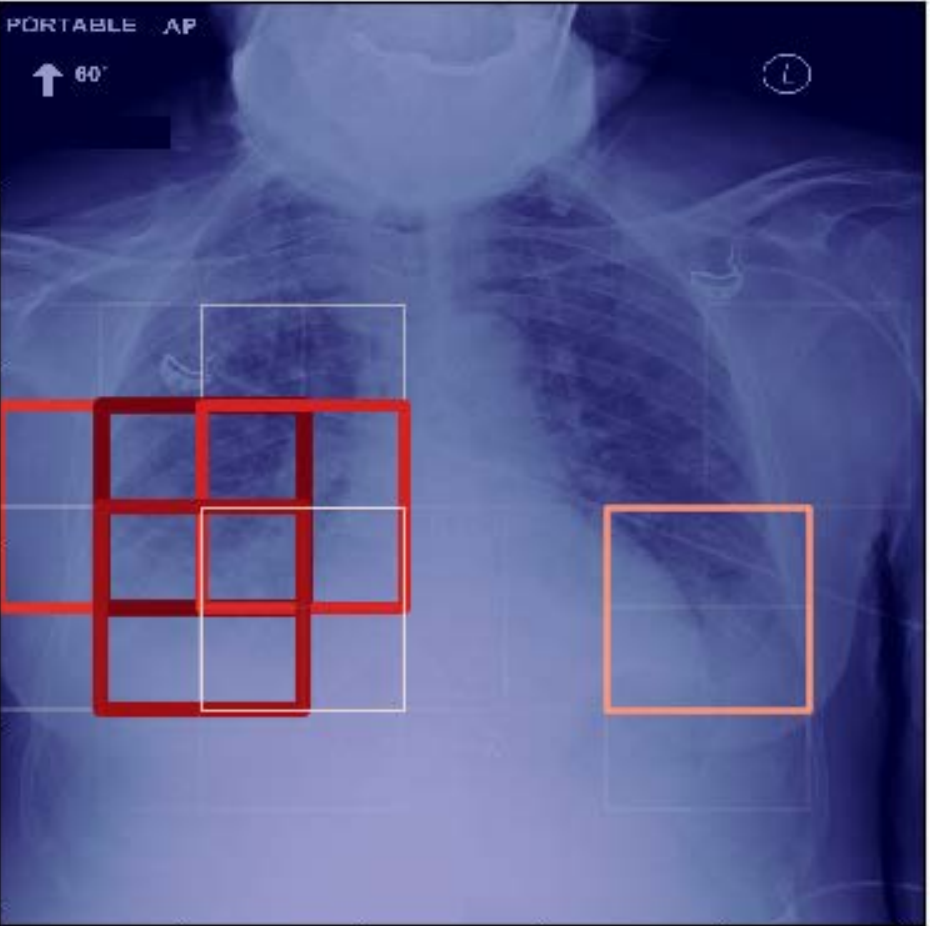}
     \includegraphics[width=0.15\textwidth, trim={0 0 0 0}, clip]{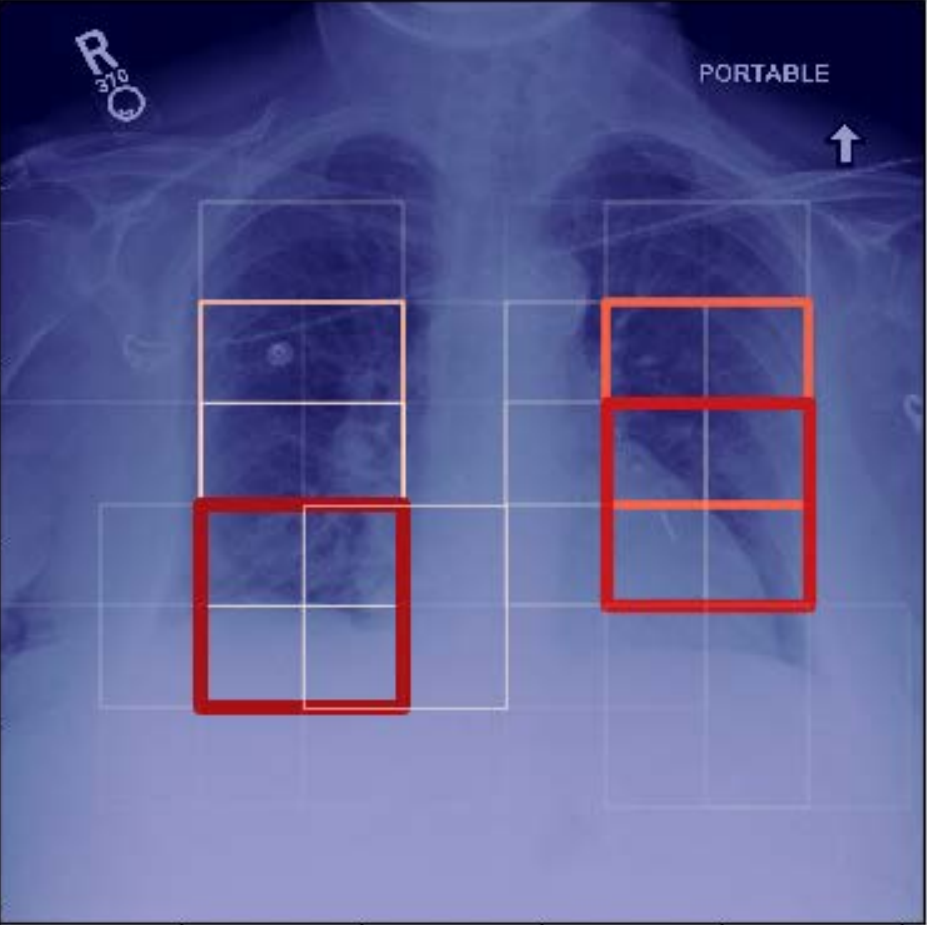}
     \includegraphics[width=0.15\textwidth, trim={0 0 0 0}, clip]{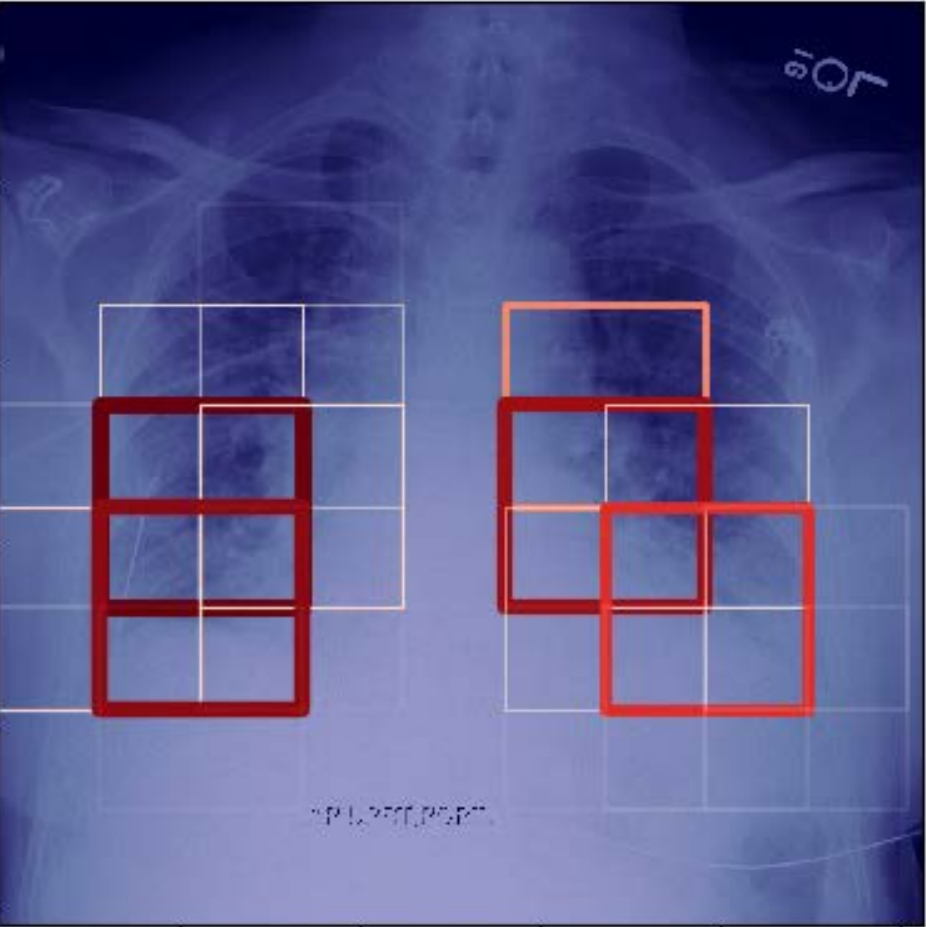}
   \caption{MIL localization results for PE with no ground truth annotation. Results are consistent in lower lungs.}
   \label{fig:resultsPE}
         \vspace{-10pt}
\end{figure}

We show critical finding localization results for several cases of PTX in Fig.~\ref{fig:resultsPTX}, PNA in Fig.~\ref{fig:resultsPNA} and PE in Fig.~\ref{fig:resultsPE}. Our visualization can be described as follows: each patch in an image has a predicted score between 0 and 1 of containing a critical finding with a correlated patch border color and line thickness. Patches close to 1 will be thick and dark red, patches with mid-range score will appear light red,  and patches closer to 0 will be thin and white. Patches that are nearly 0 will appear absent from the image. 

In Fig.~\ref{fig:resultsPTX}, ground truth PTX segmentations are highlighted in a transparent red overlay.  We notice that our method is capable of correctly localizing critical findings of various shapes, sizes, and locations, and can identify multiple findings in an image (see row 1, col 5 and row 2, col 2.). Row 3, cols 4 and 5 show images correctly classified with no PTX since all patches are close to 0. In Fig.~\ref{fig:resultsPNA}, we also notice the ability to correctly identify multiple occurrences of PNA in an image. In Fig.~\ref{fig:resultsPE}, while PE has no ground truth localization, we see consistent localization results in the lower lungs.

\section{Conclusion}
\label{sec:conclusion}
We have presented a new MIL algorithm to jointly classify and localize critical findings in CXR with competitive classification accuracy compared to state-of-the-art methods. Our MIL framework provides localizations as an interpretable explanation for the classification and does not need expensive local annotations for training.  This means we can rapidly scale to any number of critical findings much faster than with methods that require local annotations for each one. In addition, because the proposed method is built on patch-based CNNs, future applications can extend this to multi-class patch-based localization, patch-based regression for severity estimation, and additional patch-based heatmap saliency for finer localization detail.




\bibliographystyle{IEEEbib}
\bibliography{mil18}

\begin{thebibliography}{10}

\bibitem{larson2014actionable}
Paul~A. Larson, Lincoln~L. Berland, Brent Griffith, Charles~E. Kahn~Jr., and
  Lawrence~A. Liebscher,
\newblock ``Actionable findings and the role of it support: report of the {ACR}
  actionable reporting work group,''
\newblock {\em Journal of the American College of Radiology}, vol. 11, no. 6,
  pp. 552--558, 2014.

\bibitem{selvaraju2017grad}
Ramprasaath~R. Selvaraju et~al.,
\newblock ``Grad-cam: Visual explanations from deep networks via gradient-based
  localization.,''
\newblock in {\em International Conference in Computer Vision (ICCV)}, 2017,
  pp. 618--626.

\bibitem{wang2017chestx}
Xiaosong Wang, Yifan Peng, Le~Lu, Zhiyong Lu, Mohammadhadi Bagheri, and
  Ronald~M. Summers,
\newblock ``Chestx-ray8: Hospital-scale chest x-ray database and benchmarks on
  weakly-supervised classification and localization of common thorax
  diseases,''
\newblock in {\em Computer Vision and Pattern Recognition (CVPR)}. IEEE, 2017,
  pp. 3462--3471.

\bibitem{guan2018diagnose}
Qingji Guan, Yaping Huang, Zhun Zhong, Zhedong Zheng, Liang Zheng, and Yi~Yang,
\newblock ``Diagnose like a radiologist: Attention guided convolutional neural
  network for thorax disease classification,''
\newblock {\em arXiv preprint arXiv:1801.09927}, 2018.

\bibitem{redmon2016you}
Joseph Redmon, Santosh Divvala, Ross Girshick, and Ali Farhadi,
\newblock ``You only look once: Unified, real-time object detection,''
\newblock in {\em Computer Vision and Pattern Recognition (CVPR)}, 2016, pp.
  779--788.

\bibitem{li2018thoracic}
Zhe Li et~al.,
\newblock ``Thoracic disease identification and localization with limited
  supervision,''
\newblock in {\em Proceedings of the IEEE Conference on Computer Vision and
  Pattern Recognition}, 2018, pp. 8290--8299.

\bibitem{ronneberger2015u}
Olaf Ronneberger, Philipp Fischer, and Thomas Brox,
\newblock ``{U-Net}: Convolutional networks for biomedical image
  segmentation,''
\newblock in {\em International Conference on Medical Image Computing and
  Computer-Assisted Intervention (MICCAI)}. Springer, 2015, pp. 234--241.

\bibitem{yan2016multi}
Zhennan Yan et~al.,
\newblock ``Multi-instance deep learning: Discover discriminative local
  anatomies for bodypart recognition,''
\newblock {\em IEEE Transactions on Medical Imaging}, vol. 35, no. 5, pp.
  1332--1343, 2016.

\bibitem{zhu2017deep}
Wentao Zhu, Qi~Lou, Yeeleng~Scott Vang, and Xiaohui Xie,
\newblock ``Deep multi-instance networks with sparse label assignment for whole
  mammogram classification,''
\newblock in {\em International Conference on Medical Image Computing and
  Computer-Assisted Intervention (MICCAI)}. Springer, 2017, pp. 603--611.

\bibitem{xu2014deep}
Yan Xu et~al.,
\newblock ``Deep learning of feature representation with multiple instance
  learning for medical image analysis,''
\newblock in {\em Acoustics, Speech and Signal Processing (ICASSP)}. IEEE,
  2014, pp. 1626--1630.

\bibitem{kraus2016classifying}
Oren~Z Kraus, Jimmy~Lei Ba, and Brendan~J Frey,
\newblock ``Classifying and segmenting microscopy images with deep multiple
  instance learning,''
\newblock {\em Bioinformatics}, vol. 32, no. 12, pp. i52--i59, 2016.

\bibitem{johnson2019mimic}
Alistair~E.W. Johnson et~al.,
\newblock ``Mimic-cxr: A large publicly available database of labeled chest
  radiographs,''
\newblock {\em arXiv preprint arXiv:1901.07042}, 2019.

\bibitem{GoossenMIDL2019}
Andr\'{e} Goo{\ss}en, Hrishikesh Deshpande, Tim Harder, Evan Schwab, Ivo
  Baltruschat, Thusitha Mabotuwana, Nathan Cross, and Axel Saalbach,
\newblock ``Pneumothorax detection and localization in chest radiographs: A
  comparison of deep learning approaches,''
\newblock in {\em Medical Imaging with Deep Learning (MIDL)}, 2019.

\bibitem{rajpurkar2017chexnet}
Pranav Rajpurkar et~al.,
\newblock ``Chexnet: Radiologist-level pneumonia detection on chest {X}-rays
  with deep learning,''
\newblock {\em arXiv preprint arXiv:1711.05225}, 2017.

\end{thebibliography}

\end{document}